\pdfoutput=1  % arXiv: force the pdflatex route

\documentclass[10pt,twocolumn,letterpaper]{article}

\usepackage{cvpr}                % CAMERA-READY version
\usepackage{multirow}
\usepackage{pifont}   % \ding{41} = envelope, marks the corresponding author

\makeatletter
\renewcommand\paragraph{\@startsection{paragraph}{4}{\z@}%
  {1.3ex \@plus .25ex \@minus .2ex}%
  {-1em}%
  {\normalfont\normalsize\bfseries}}
\makeatother
\definecolor{cvprblue}{rgb}{0.21,0.49,0.74}
\usepackage[breaklinks,colorlinks,allcolors=cvprblue]{hyperref}

\def\paperID{0000}
\def\confName{CVPR}
\def\confYear{2026}

\newcommand{\method}{ACE-Ego-Hand} % paper-facing name (internal codename: OffWan)
\newcommand{\best}[1]{\textbf{#1}}
\definecolor{teaseryellow}{rgb}{0.85,0.68,0.0}

\title{\method{}: Repurposing Video Diffusion Models for Occlusion-Robust Egocentric 3D Hand Motion Recovery}

\author{%
Yufei Liu\textsuperscript{1,4}, Xixi Wang\textsuperscript{2}, Hao Li\textsuperscript{3,4}, Ganlong Zhao\textsuperscript{3,4}, Kaitong Cai\textsuperscript{4}, Chengkai Jin\textsuperscript{2,4},\\
Chunxiao Liu\textsuperscript{4}, Jianbo Liu\textsuperscript{4}, Siyuan Huang\textsuperscript{4\,\dag}, Xingang Pan\textsuperscript{2}, Hongsheng Li\textsuperscript{3,4\,\ding{41}}\\[0.5em]
\textsuperscript{1}Shanghai Jiao Tong University \quad \textsuperscript{2}Nanyang Technological University\\
\textsuperscript{3}The Chinese University of Hong Kong \quad \textsuperscript{4}ACE Robotics\\[0.35em]
{\small\dag\,Project leader.\quad\ding{41}\,Corresponding author.}
}

\begin{document}

% CVPR teaser idiom: \maketitle and the page-1 figure are typeset together in the
% one-column title block, so the figure lands on page 1 (a figure* never can).
\twocolumn[{%
\renewcommand\twocolumn[1][]{#1}%
\maketitle
\centering
    \includegraphics[width=\linewidth]{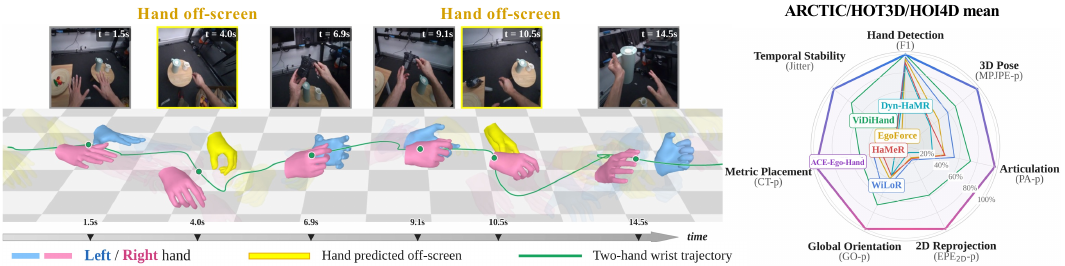}
    \captionof{figure}{\textbf{Occlusion-robust egocentric 3D hand motion recovery.} Left: \method{} applied to a raw RGB video clip. \method{} maintains hand identity and trajectory continuity even when a hand temporarily leaves the field of view (frames shaded \textcolor{teaseryellow}{yellow}). Right: \method{} leads on six of the seven metrics shown (F1, MPJPE-p, PA-p, EPE\textsubscript{2D}-p, GO-p, CT-p, and Jitter, each averaged over ARCTIC, HOT3D, and HOI4D, with farther from the center being better) and matches the strongest baseline on detection F1.}
    \label{fig:teaser}
    \vspace{0.5em}
}]

\begin{abstract}
Egocentric video offers scalable manipulation data for embodied AI, yet recovering metric 3D hand trajectories remains challenging due to severe object occlusion and frequent out-of-sight gaps. Existing single-frame and windowed temporal regressors fail when a hand briefly leaves the frame, while recent video diffusion models (VDMs) rely on heavy, stochastic multi-step sampling as pixel-space renderers.
We instead repurpose a VDM into a deterministic geometry encoder. A single forward pass over the clean latent exposes scene content beyond current observations, including occluded and out-of-sight hands. We introduce \textbf{\method{}}, an offline clip-level framework that extracts features via a Deterministic Clean-Latent Encoder and decodes them with a Bidirectional Spatiotemporal Decoder. \method{} recovers continuous bimanual trajectories with metric placement and no external detector, while a Ray-Based Camera Solver supports a second configuration that needs no test-time camera intrinsics.
Across five egocentric benchmarks, \method{} sets a new state of the art, cutting MPJPE-p by $30\%$ on occlusion-heavy ARCTIC and $40\%$ on HOT3D. These gains reach $46\%\text{--}61\%$ once out-of-sight hands are included in the evaluation, offering a scalable path from everyday human video to robot manipulation data. The code is publicly available at \href{https://github.com/ggxxii/ACE-Ego-Hand}{github.com/ggxxii/ACE-Ego-Hand}.
\end{abstract}

\section{Introduction}
Egocentric video offers a scalable source of robot manipulation data~\citep{egomimic2024,egovla2025,egodex2025,li2026aceego0unifyingegocentrichuman}, but exploiting this source requires recovering metric 3D hand motion from raw footage. 
Two constraints of egocentric capture make this recovery particularly challenging: frequent hand-object-interaction (HOI) occlusions, and swift head motion that pushes hands entirely out of sight (OOS) and leaves no visual evidence in the field of view (Figure~\ref{fig:teaser}).
Existing 3D hand motion reconstruction methods rely heavily on per-frame detectors~\citep{hamer2024,wilor2025,interwild2023,wildhands2024,hamba2024} or SLAM-anchored tracking~\citep{hawor2025,dynhamr2025,tram2024,slahmr2023}, the latter inherently suffering from error accumulation. Once a hand becomes unobserved or exits the camera frustum, windowed temporal models~\citep{omnihands2025,tcmr2021,glot2023} lose historical context and struggle to resume coherent tracking upon re-emergence. 
To tackle this, we argue that metric 3D hand trajectory recovery from egocentric video is fundamentally an offline clip-level spatiotemporal reasoning task.

In order to effectively reconstruct plausible hand dynamics during OOS intervals, rich physical and temporal priors are essential. 
Pretrained video diffusion models (VDMs) naturally capture such priors of physical dynamics and human-object interaction.  
Recent generative frameworks~\cite{vidihand2026} exploit video diffusion priors for hand pose estimation under severe HOI occlusions. 
However, they rely on an indirect pixel-rendering pipeline—executing multi-step sampling to synthesize pixel-space outputs from which poses are then regressed. 
This roundabout route incurs sampling latency and may propagate generative artifacts into the pose estimates. 
Furthermore, directly regressing 3D hand coordinates in the camera frame conflates hand articulation with rapid camera ego-motion and imposes strict calibration dependencies. 

To address these limitations, we present \textbf{\method{}}, an offline clip-level 3D hand motion reconstruction framework.
Our key insight is twofold: the VDM serves reconstruction best as an encoder rather than a renderer, and its generic features need end-to-end 3D supervision to become geometry-aware. \method{} therefore repurposes a pretrained VDM~\citep{wan2025} as a Deterministic Clean-Latent Encoder. 
A single noise-free forward pass over the full video clip yields a dense spatiotemporal feature grid, adapted end-to-end via LoRA to capture structured physical representations without generative sampling latency.
To recover continuous metric trajectories from these features, we propose a Bidirectional Spatiotemporal Decoder paired with a Ray-Based Camera Solver. 
Across the entire video clip, we aggregate long-range temporal context via unconstrained bidirectional attention. 
By anchoring joint-specific queries to spatial feature maps and enforcing a global shape prior, the model reliably interpolates plausible hand poses during severe occlusions and prolonged out-of-sight intervals.
Concurrently, to decouple hand placement from camera intrinsics and rapid ego-motion, we estimate a 3D viewing-ray field directly from the diffusion latents. 
By optimizing translation against predicted 2D anchors during visible frames and relying on temporal bearing cues during visual gaps, \method{} can recover metric hand translation without a supplied calibration matrix, which supports intrinsics-free ($K$-free) inference at test time.

Our primary contributions are summarized as follows:
\begin{itemize}
\item We introduce \method{}, an offline clip-level framework that repurposes a video diffusion model into a Deterministic Clean-Latent Encoder, recovering metric bimanual trajectories without any external detector.
\item We present a Bidirectional Spatiotemporal Decoder that synergizes spatially grounded queries, clip-level shape priors, and bidirectional reasoning to reconstruct out-of-sight hands. A Ray-Based Camera Solver further decouples hand motion from camera dynamics to enable test-time intrinsics-free inference.
\item Extensive evaluations on five challenging egocentric benchmarks demonstrate that \method{} achieves superior accuracy and robustness, outperforming state-of-the-art baselines on $45$ of $48$ primary metric comparisons and remaining stable under severe occlusions and hand absences.
\end{itemize}

\section{Related Work}

\paragraph{Single-frame hand reconstruction.}
Most single-frame methods follow a ``detect--crop--regress'' pipeline, localizing the hand and then regressing MANO parameters from the cropped image~\citep{hamer2024,wilor2025,hamba2024,interwild2023}. For egocentric settings, WildHands~\citep{wildhands2024} conditions on camera intrinsics, and EgoForce~\citep{egoforce2026} adds a forearm-guided camera-space solver with causal translation filtering at inference. Despite the improved camera modeling, these methods remain crop-dependent. Scale and depth are tied to the detection box, and they produce no reconstruction when the hand is severely occluded or outside the field of view.

\paragraph{Video-based hand motion recovery.}
Video-based methods address missing observations through temporal fusion or motion-prior completion. Early works apply bidirectional modeling to per-frame features~\citep{vibe2020} or aggregate multiple frames to predict the center frame~\citep{tcmr2021,glot2023}, and OmniHands~\citep{omnihands2025} fuses nine consecutive two-hand crops to recover hands fully occluded \emph{within} the image. A second line completes invisible hands with motion priors. GENMO~\citep{genmo2025} performs bidirectional motion modeling conditioned on video features, and EgoH4~\citep{egoh4_2025} causally predicts out-of-sight hands from estimated body poses. However, their visual front ends still rely on frame-wise detection and cropping, and they typically complete motion over normalized pose sequences rather than the original video~\citep{mdm2023}.

\paragraph{Generative video models as representations.}
Recent work increasingly repurposes generative video models for downstream perception, either by reading their activations~\citep{genception2026} or by prompting the generation process itself~\citep{pointprompting2026}. The two routes differ in cost. Reading activations needs a single forward pass, while prompting the generation process pays for a full sampling trajectory. Self-supervised video encoders~\citep{vjepa2_2025,videomae2022} offer a discriminative alternative, trained to predict masked content rather than to synthesize it. ViDiHand~\citep{vidihand2026} applies the generation-prompting route to hand reconstruction. The method fine-tunes Wan through VACE~\citep{vace2025} to synthesize geometric overlays and regresses MANO from intermediate denoising features, using $12$--$25$ sampling steps and a decoder trained on cached frozen features. The backbone therefore receives no gradient from the reconstruction objective, and those features stay generic. We instead read the clean latent in a single forward pass inside the training loop.

\begin{figure*}[t]
\centering
\includegraphics[width=1.0\textwidth]{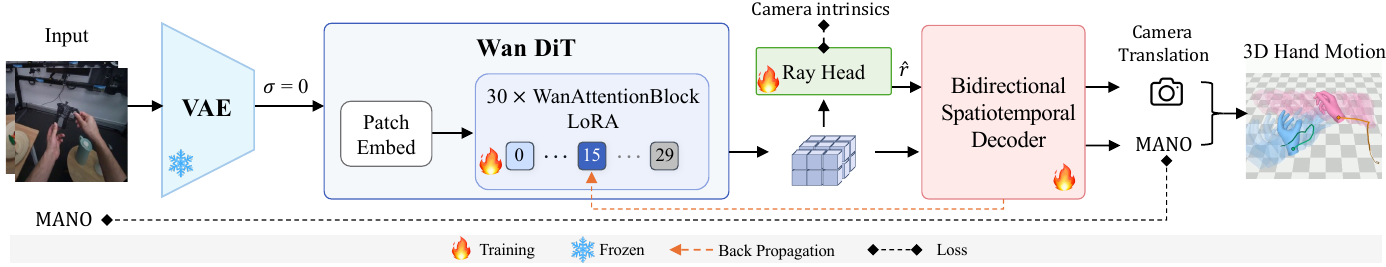}
\caption{\textbf{Overview of \method{}.} Raw video is encoded into latents by the Wan VAE and processed by the Wan DiT ($\sigma=0$). Block-15 features (grayed blocks skipped) branch to the Ray Head and Bidirectional Spatiotemporal Decoder. Only the patch embedding, LoRA, Ray Head, and Decoder are trained (parameter accounting in Appendix~\ref{sec:supp_impl}). Camera intrinsics supervise the Ray Head and the $K$-free camera fit during training. At test time they are discarded ($K$-free) or used only as the bearing source in the translation solve (standard).}
\label{fig:method}
\vspace{-1.0em}
\end{figure*}

\section{\method{}}
\label{sec:method}

\method{} processes an egocentric video clip $V=\{I_t\}_{t=1}^{T}$ to predict frame-wise 3D bimanual MANO parameters in a single deterministic pass: global orientation $\hat R_t$, articulation $\hat\theta_t$, camera-frame translation $\hat\tau_t$, and a per-clip shape $\hat\beta$. The MANO layer $\mathcal{M}$ maps the pose and shape parameters to hand meshes and $21$ joints, and the network additionally predicts per-frame existence and visibility flags. Instead of iteratively sampling from generative models, we extract motion representations directly through three unified modules (Figure~\ref{fig:method}): a \textbf{Deterministic Clean-Latent Encoder} that reads features from a LoRA-adapted pretrained video diffusion model, a \textbf{Bidirectional Spatiotemporal Decoder} for sequence-wide trajectory estimation, and a \textbf{Ray-Based Camera Solver}. We consider two configurations: standard \method{} and the intrinsics-free \method{}.

\begin{figure}[t]
\centering
\includegraphics[width=1.0\columnwidth]{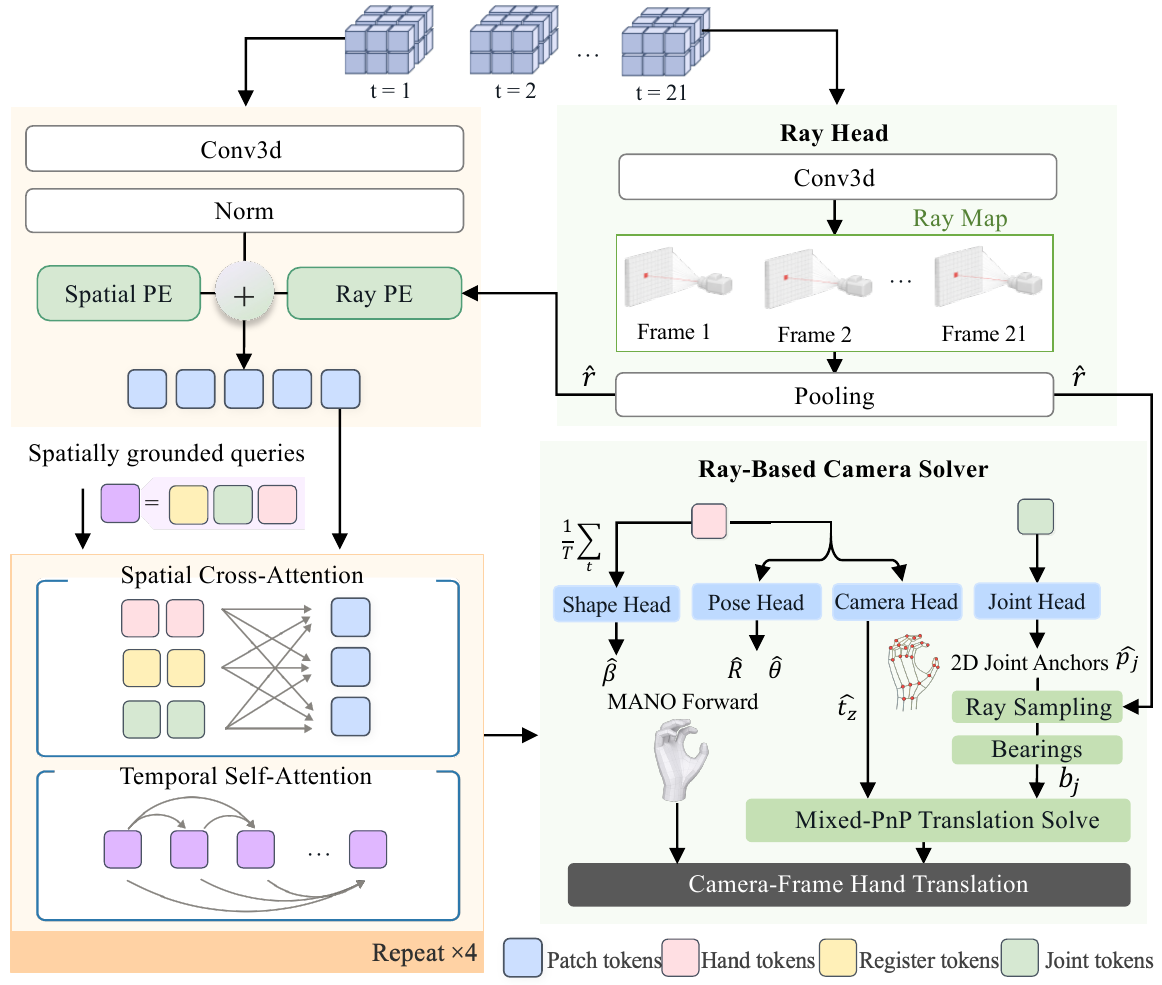}
\caption{\textbf{Architecture of the Bidirectional Spatiotemporal Decoder.} The predicted ray field $\hat r$ enters as a ray PE added to the spatial PE. Four layers alternate Spatial Cross-Attention with Temporal Self-Attention. The readout heads emit MANO parameters and feed the mixed-PnP translation solve. Our two configurations differ architecturally only in the bearing source, derived from the predicted ray field $\hat r$ ($K$-free, shown) or computed from camera intrinsics (standard).}
\label{fig:projector}
\vspace{-1.0em}
\end{figure}

\subsection{From Generator to Encoder}
\paragraph{Feedforward encoding.}
The generator $\Phi$ is a Diffusion Transformer (DiT) pretrained via rectified flow on noisy latents $x_\sigma = (1{-}\sigma)z + \sigma\epsilon$, where $z$ is the clean video latent, $\epsilon$ is Gaussian noise, and $\sigma\in[0,1]$ is the noise level. This sequence-level pretraining is expected to equip $\Phi$ with spatiotemporal priors such as object permanence, 3D structural consistency, and occlusion reasoning. \cref{sec:ablation} probes this premise by swapping feature sources. We therefore use $\Phi$ strictly as an encoder. A frozen VAE encoder $\mathcal{E}$ compresses the clip into the clean latent $z=\mathcal{E}(V)$, and a single deterministic forward pass runs at zero noise ($\sigma=0$):
\begin{equation}
F = \Phi_{0:L^\star}(z; \sigma=0),
\label{eq:encoder}
\end{equation}
where $\Phi_{0:L^\star}$ truncates execution at block $L^\star=15$ of the zero-indexed $30$-block stack, running the first $16$ blocks. Bypassing the remaining blocks and generation head cuts per-pass computation by approximately half. A tap-depth sweep in Appendix~\ref{sec:supp_analysis} shows that block~$15$ retains nearly all of the available accuracy.
The resulting feature sequence $F=\{F_\ell\}_{\ell=1}^{T'}$ contains $T'=21$ latent frames for an input of $T=81$ frames. This setup yields $4\times$ temporal and $16\times$ spatial compression. Each latent frame $F_\ell$ forms a spatial grid of $D=3072$-channel feature cells. For a $672\times 480$ input, this corresponds to a $42\times 30$ grid of $16\times 16$ pixel patches.
The latent features $F$ then feed both the Bidirectional Spatiotemporal Decoder (\cref{sec:decoder}) and the Ray-Based Camera Solver (\cref{sec:camera}).

\paragraph{End-to-end adaptation.}
We keep the encoder trainable within the optimization loop, utilizing LoRA~\citep{lora2022} on attention and feed-forward projections alongside a trainable patch embedding layer, so that backpropagated 3D supervision reshapes $F$ into geometry-aware features end to end. We compare alternative latent representations in \cref{sec:ablation}. A matched comparison in Appendix~\ref{sec:supp_analysis} further shows that reading the clean latent at $\sigma=0$ outperforms reading noised latents.

\subsection{Bidirectional Spatiotemporal Decoder}
\label{sec:decoder}

\paragraph{Spatially grounded queries.}
We propose a lightweight Bidirectional Spatiotemporal Decoder with spatially grounded queries to extract representations from the tapped feature grid, as illustrated in Figure~\ref{fig:projector}. The decoder processes $48$ queries per latent frame: $2$ hand tokens assigned to fixed left and right slots, $42$ joint tokens, and $4$ register tokens that serve as learned scratch space without updating $F$. Structuring joint queries as spatially grounded tokens binds features directly to physical keypoints, facilitating precise 3D hand pose estimation.

Each latent frame is tokenized into patch tokens by integrating two additive positional encodings:
\begin{equation}
X_\ell = \mathrm{LN}(W_F F_\ell) + P^{\mathrm{sp}} + g\big(\Gamma(\hat r)\big),
\label{eq:token}
\end{equation}
where $\mathrm{LN}$ denotes Layer Normalization and $W_F$ projects the $D=3072$ backbone channels to the 384-dimensional decoder space. The spatial PE $P^{\mathrm{sp}}$ provides explicit coordinates for attention, and the ray PE injects viewing geometry: $\Gamma$ Fourier-encodes the direction of each cell's predicted viewing ray $\hat r$, produced by the Ray Head (\cref{sec:camera}), and a zero-initialized MLP $g$ maps that encoding to the decoder dimension (implementation details in Appendix~\ref{sec:supp_impl}).

\paragraph{Spatial readout heads.}
The Joint Head estimates 2D joint locations directly from this spatial grid. For each joint $j$, the cross-attention weights from its corresponding token form a spatial heatmap $A_j$. A soft-argmax operation then aggregates these attention probabilities to derive 2D joint anchors $\hat p_j$:
\begin{equation}
\hat p_j = \sum_{u} A_j(u)\,u, \qquad \sum_{u} A_j(u)=1,
\label{eq:softargmax}
\end{equation}
where $u$ iterates over normalized grid-cell center coordinates in $[0,1]^2$. This differentiable formulation anchors each keypoint with sub-cell precision, and an MLP then predicts wrist-relative 3D joint positions in meters. In parallel, the Pose Head regresses the global orientation $\hat R$ and articulation $\hat\theta$ as Gram--Schmidt-orthogonalized 6D rotations, and the Camera Head predicts a log-depth $\hat\zeta$ with $\hat t_z = \exp(\hat\zeta)$ guaranteeing strictly positive metric depth. Existence and visibility confidence scores complete the frame-level readout, avoiding the need for Hungarian matching~\citep{kuhn1955,detr2020}.

\paragraph{Clip-level shape prior.}
In egocentric videos, an individual hand maintains a constant physical shape and scale throughout a continuous recording. Estimating mesh parameters independently per frame, however, risks size flickering and shape drift under camera motion and local occlusions. To enforce this physical invariant, the Shape Head predicts hand shape parameters $\hat\beta$ once per hand per video clip by temporally pooling hand tokens across all $T$ frames, yielding a single mesh scale for the entire sequence.

\paragraph{Unconstrained bidirectional reasoning.}
We process the spatially grounded queries with four alternating attention layers. Frame $\ell$'s queries first extract per-frame visual details through Spatial Cross-Attention over $X_\ell$, and queries from all frames then exchange motion information via Temporal Self-Attention. Rotary relative positional encodings~\citep{rope2024} on the temporal axis remove absolute sequence constraints, so a single forward pass generalizes to long recordings (Appendix~\ref{sec:supp_analysis}). Temporal attention runs at the downsampled latent frame rate $T'$, with outputs linearly interpolated back to the $T$ video frames, reducing whole-clip attention to roughly $(T'/T)^2 \approx 1/15$ of the full-resolution cost. We apply no causal mask: each frame conditions on both past and future context across the entire clip, so the decoder reconstructs occluded or out-of-sight hands rather than extrapolating unidirectionally.

\subsection{Ray-Based Camera Solver}
\label{sec:camera}
\paragraph{Intrinsics-free ray field prediction.}
Camera geometry is needed at two points: the ray positional encoding in Eq.~\eqref{eq:token} and the metric projection of the predicted hand. Readout heads that memorize the pixel-to-metric mapping of one camera fail to generalize across intrinsics, so we instead predict a continuous viewing-ray field, following ray-based camera representations~\citep{raydiffusion2024,perspectivefields2023}. The Ray Head, a zero-initialized $1\times 1$ convolution on $F$, predicts a per-cell direction normalized to a unit ray $\hat r = (\hat r_x, \hat r_y, \hat r_z)$ in the camera frame. Temporal pooling then averages the per-frame predictions into a single field, since intrinsics are constant within a clip.

During training, ground-truth camera calibration supervises this ray field using a cosine distance loss:
\begin{equation}
\mathcal{L}_{\mathrm{ray}} = \frac{1}{|\Omega|} \sum_{u\in\Omega} \big(1 - \langle \hat r(u),\, r_K(u) \rangle\big),
\label{eq:ray}
\end{equation}
where $\langle\cdot,\cdot\rangle$ denotes the inner product, $\Omega$ is the latent token grid, and $r_K(u)$ is the unit ray unprojected from the cell center under the calibrated camera model. One head thus accommodates both pinhole and fisheye camera models without intrinsics as input at test time.

\paragraph{Mixed-PnP translation.}
To determine metric placement, we employ a mixed Perspective-n-Point (PnP) scheme, similar in spirit to \citet{vidihand2026} but formulated for the calibration-free setting. Rather than predicting full 3D translations, the Camera Head regresses only the optical depth $\hat t_z = \exp(\hat\zeta)$, and we solve the in-plane translation $(t_x, t_y)$ directly against the predicted 2D joint anchors. Per frame and hand (indices suppressed), the MANO forward pass yields $J^{\mathrm{can}} = \mathcal{M}(\hat R, \hat\theta, \hat\beta)$, the $21$ joints posed and camera-oriented but not yet placed, so the depth of joint $j$ along the optical axis is $z_j = J^{\mathrm{can}}_{z,j} + \hat t_z$. The bearing vector $b_j = (b^x_j, b^y_j)$, the dimensionless pair $(x/z, y/z)$ of the ray toward joint $j$, is evaluated analytically from the predicted ray field $\hat r$. A closed-form per-axis regression fits an effective pinhole camera $(\hat f, \hat c)$ to $\hat r$ in normalized pixel units, with no calibration input, and $b_j = (\hat p_j - \hat c)/\hat f$. The fitted camera extrapolates to out-of-frame anchors and denoises the per-token field. Standard \method{} instead computes the bearings from the provided intrinsics as $\big((u_j{-}c_x)/f_x,\,(v_j{-}c_y)/f_y\big)$ with $(u_j, v_j) = \hat p_j$ in pixels. Each configuration is trained end to end with its own bearing source, the $K$-free rows in Table~\ref{tab:main} report a separately trained model, not a test-time solver switch.
% with one added supervision term for the $K$-free configuration (\cref{sec:recipe}), so 

Perspective projection is linear in the in-plane shift, so $t_x$ admits a closed-form weighted least-squares solution:
\begin{equation}
b^x_j = \frac{J^{\mathrm{can}}_{x,j} + t_x}{z_j} 
\quad \Rightarrow \quad 
\hat t_x = \frac{\sum_j m_j z_j^{-1} \big(b^x_j - J^{\mathrm{can}}_{x,j} / z_j\big)}{\sum_j m_j z_j^{-2}},
\label{eq:pnp}
\end{equation}
where $m_j \in \{0, 1\}$ selects joints that lie in front of the camera with anchors inside the frame. We solve for $\hat t_y$ symmetrically, yielding $\hat\tau = (\hat t_x, \hat t_y, \hat t_z)$. With too few valid joints or an excessive re-projection residual, the hand falls back to its inverse-projected wrist ray at depth $\hat t_z$ (thresholds in Appendix~\ref{sec:supp_impl}).

\subsection{Training Recipe}
\label{sec:recipe}
We optimize all trainable components, namely the patch embedding, the LoRA modules, the Ray Head, and the Bidirectional Spatiotemporal Decoder, jointly under a unified objective:
\begin{equation}
\mathcal{L} = \mathcal{L}_{\mathrm{rot}} + \mathcal{L}_{\mathrm{joint}} + \mathcal{L}_{\mathrm{img}} + \mathcal{L}_{\mathrm{cam}} + \mathcal{L}_{\mathrm{pres}} + \mathcal{L}_{\mathrm{tmp}} + \mathcal{L}_{\mathrm{ray}}.
\label{eq:loss}
\end{equation}
$\mathcal{L}_{\mathrm{rot}}$ supervises orientation, articulation, and shape. $\mathcal{L}_{\mathrm{joint}}$ constrains root-relative, camera-frame, and wrist 3D positions. $\mathcal{L}_{\mathrm{img}}$ penalizes 2D anchors and re-projected MANO keypoints under the training camera. $\mathcal{L}_{\mathrm{cam}}$ supervises camera-frame translation, with gradients flowing through the PnP solver. $\mathcal{L}_{\mathrm{pres}}$ trains existence and visibility scores, and $\mathcal{L}_{\mathrm{tmp}}$ penalizes 3D joint accelerations for temporal smoothness. The $K$-free configuration adds $\mathcal{L}_{\mathrm{fit}}$, the bearing error of its learned camera (\cref{sec:camera}) against the calibrated one, whose gradient reaches the ray field only through the four fitted parameters. Loss weights are in Appendix~\ref{sec:supp_impl}.

We highlight three key supervision strategies. First, we fully supervise out-of-sight hands rather than masking them out, forcing the bidirectional attention to reconstruct invisible hands from temporal context. Second, a source lacking 3D MANO annotations, RHD in our mixture, supervises only 2D anchors, 3D joints, and presence heads, a routing that adds appearance diversity while shielding the MANO parameter heads. Third, camera intrinsics never enter the encoder or decoder: they serve as training targets for the $\mathcal{L}_{\mathrm{img}}$ re-projections, $\mathcal{L}_{\mathrm{ray}}$, and $\mathcal{L}_{\mathrm{fit}}$, and in the standard configuration, they additionally supply the bearings of the translation solve.

\section{Experiments}
\label{sec:exp}
\subsection{Experimental Setup}

\paragraph{Training data and benchmarks.}
Both \method{} configurations are trained separately for $20$k steps, the standard one on $16$ A100 GPUs and the $K$-free one on $8$, using a weighted mixture of MANO-labeled egocentric video (ARCTIC, HOT3D, H2O, and OakInk2), rendered egocentric video from Re:InterHand~\citep{reinterhand2023}, and two image-only hand datasets (FreiHAND~\citep{freihand2019} and RHD~\citep{rhd2017}). HOI4D serves as the held-out test set.

We evaluate \method{} across five complementary benchmarks: ARCTIC~\citep{arctic2023} for severe bimanual occlusion, HOT3D~\citep{hot3d2024} for complex egocentric viewpoints, HOI4D~\citep{hoi4d2022} for category-level 4D interaction generalization, H2O~\citep{h2o2021} for bimanual hand-object pose estimation, and OakInk2~\citep{oakink2_2024} for long-horizon multi-object manipulation. The main text reports ARCTIC, HOT3D, and held-out HOI4D. H2O and OakInk2 appear in Appendix~\ref{sec:supp_results}, and the splits and preprocessing in Appendix~\ref{sec:supp_protocol}.

\begin{table*}[t]
\centering
\setlength{\tabcolsep}{3.2pt}
\renewcommand{\arraystretch}{0.79}
\caption{\textbf{Main comparison on ARCTIC, HOT3D, and held-out HOI4D.} Best result per column in bold. $\uparrow$/$\downarrow$ give the direction of improvement. MPJPE\textsuperscript{+OOS} averages over \emph{all} hand-frames, in view and out of sight alike, so it needs out-of-sight ground truth, which HOI4D lacks (--). $^{\dagger}$ zero-shot evaluation, $^{\ddagger}$ causal Kalman filtering.}
\label{tab:main}
\resizebox{\textwidth}{!}{%
\begin{tabular}{@{}ll ccc ccc ccc c@{}}
\toprule
& & \multicolumn{3}{c}{Detection} & \multicolumn{3}{c}{3D Pose} & \multicolumn{3}{c}{Orient.\ \& Position} & Temporal \\
\cmidrule(lr){3-5}\cmidrule(lr){6-8}\cmidrule(lr){9-11}\cmidrule(lr){12-12}
& Method & FAcc$\uparrow$ & Recall$\uparrow$ & F1$\uparrow$ & MPJPE-p$\downarrow$ & PA-p$\downarrow$ & MPJPE\textsuperscript{+OOS}$\downarrow$ & EPE\textsubscript{2D}-p$\downarrow$ & GO-p$\downarrow$ & CT-p$\downarrow$ & Jitter$\downarrow$ \\
\midrule
\multirow{12}{*}{\rotatebox{90}{ARCTIC}}
 & InterWild & 0.878 & 0.943 & 0.959 & 30.817 & 15.952 & 39.435 & 53.888 & 25.386 & 0.097 & 46.577 \\
 & HaMeR & 0.875 & 0.943 & 0.957 & 29.197 & 14.596 & 38.183 & 65.289 & 24.907 & 0.095 & 18.279 \\
 & Hamba & 0.833 & 0.912 & 0.941 & 31.233 & 17.168 & 40.039 & 87.047 & 27.822 & 0.110 & 15.357 \\
 & WildHands & 0.879 & 0.946 & 0.960 & 25.704 & 13.941 & 33.915 & 50.517 & 22.320 & 0.058 & 12.972 \\
 & WiLoR & 0.919 & 0.951 & 0.974 & 22.012 & 11.873 & 31.502 & 71.527 & 17.358 & 0.075 & 24.091 \\
 & EgoForce$^{\ddagger}$ & 0.882 & 0.930 & 0.963 & 22.388 & 14.322 & 32.193 & 61.761 & 21.073 & 0.069 & 24.158 \\
 & OmniHands & 0.866 & 0.949 & 0.954 & 29.674 & 14.203 & 38.191 & 51.505 & 24.580 & 0.087 & 45.312 \\
 & Dyn-HaMR & 0.842 & 0.918 & 0.951 & 27.904 & 17.017 & 37.445 & 85.723 & 25.951 & 0.121 & 12.840 \\
 & HaWoR & 0.700 & 0.817 & 0.895 & 45.357 & 26.375 & 53.443 & 158.062 & 43.325 & 0.149 & 19.789 \\
 & ViDiHand & 0.997 & 0.999 & 0.999 & 21.668 & 9.821 & 31.045 & 12.407 & 14.642 & 0.047 & 3.183 \\
 & \textbf{\method{}} & \best{1.000} & \best{1.000} & \best{1.000} & \best{15.256} & \best{7.474} & \best{16.783} & \best{9.180} & \best{11.807} & \best{0.021} & \best{2.700} \\
 & \method{} ($K$-free) & 0.999 & \best{1.000} & \best{1.000} & 16.615 & 8.756 & 18.061 & 11.014 & 12.208 & 0.028 & 2.770 \\
\midrule
\multirow{12}{*}{\rotatebox{90}{HOT3D}}
 & InterWild & 0.669 & 0.881 & 0.868 & 77.168 & 24.811 & 89.218 & 71.482 & 58.501 & 0.213 & 101.164 \\
 & HaMeR & 0.692 & 0.904 & 0.883 & 68.314 & 21.455 & 80.264 & 59.077 & 49.636 & 0.102 & 23.632 \\
 & Hamba & 0.632 & 0.828 & 0.853 & 71.732 & 29.620 & 83.438 & 107.625 & 56.525 & 0.128 & 18.507 \\
 & WildHands & 0.655 & 0.863 & 0.844 & 52.791 & 28.946 & 60.491 & 111.438 & 53.933 & 0.157 & 22.885 \\
 & WiLoR & 0.827 & 0.897 & 0.937 & 30.966 & 19.980 & 52.014 & 72.978 & 25.746 & 0.098 & 17.976 \\
 & EgoForce$^{\ddagger}$ & 0.769 & 0.856 & 0.916 & 43.960 & 25.709 & 63.085 & 83.521 & 37.144 & 0.130 & 38.342 \\
 & OmniHands & 0.649 & 0.895 & 0.868 & 63.281 & 22.682 & 73.503 & 68.437 & 49.120 & 0.133 & 69.510 \\
 & Dyn-HaMR & 0.614 & 0.811 & 0.802 & 74.214 & 38.201 & 80.952 & 171.617 & 43.851 & 0.571 & 44.942 \\
 & HaWoR & 0.348 & 0.499 & 0.654 & 71.396 & 66.031 & 84.733 & 327.294 & 79.350 & 0.262 & 23.872 \\
 & ViDiHand & 0.948 & 0.974 & 0.983 & 21.514 & 11.383 & 44.440 & 14.953 & 15.829 & 0.040 & 3.741 \\
 & \textbf{\method{}} & \best{0.986} & 0.998 & \best{0.996} & \best{12.888} & \best{6.436} & \best{17.273} & 6.418 & \best{7.924} & \best{0.025} & \best{3.159} \\
 & \method{} ($K$-free) & 0.984 & \best{0.999} & 0.995 & 13.535 & 6.768 & 18.703 & \best{6.067} & 8.362 & 0.026 & 3.426 \\
\midrule
\multirow{12}{*}{\rotatebox{90}{HOI4D$^{\dagger}$}}
 & InterWild & 0.731 & 0.922 & 0.864 & 53.072 & 22.909 & -- & 80.549 & 41.743 & 0.228 & 98.866 \\
 & HaMeR & 0.731 & 0.923 & 0.864 & 44.481 & 21.580 & -- & 79.494 & 33.557 & 0.187 & 20.068 \\
 & Hamba & 0.710 & 0.885 & 0.849 & 47.161 & 25.924 & -- & 115.793 & 37.390 & 0.204 & 21.556 \\
 & WildHands & 0.730 & 0.924 & 0.864 & 45.623 & 23.601 & -- & 82.246 & 45.654 & 0.159 & 18.615 \\
 & WiLoR & 0.962 & 0.966 & 0.972 & 33.710 & 14.903 & -- & 41.579 & 25.527 & 0.115 & 17.449 \\
 & EgoForce$^{\ddagger}$ & 0.917 & 0.941 & 0.949 & 44.746 & 19.160 & -- & 75.821 & 38.861 & 0.126 & 83.452 \\
 & OmniHands & 0.655 & 0.937 & 0.834 & 44.255 & 18.689 & -- & 70.662 & 34.392 & 0.108 & 24.212 \\
 & Dyn-HaMR & 0.750 & 0.863 & 0.845 & 45.097 & 29.259 & -- & 144.643 & 40.176 & 0.258 & 17.947 \\
 & HaWoR & 0.869 & 0.864 & 0.919 & 47.329 & 28.851 & -- & 135.748 & 43.091 & 0.139 & 28.376 \\
 & ViDiHand & \best{0.984} & 0.991 & \best{0.990} & 30.090 & 13.960 & -- & 24.460 & 23.420 & 0.117 & 4.010 \\
 & \textbf{\method{}} & 0.958 & 0.996 & 0.974 & 23.031 & 11.664 & -- & \best{14.987} & 18.426 & 0.057 & 2.393 \\
 & \method{} ($K$-free) & 0.975 & \best{0.999} & 0.985 & \best{22.154} & \best{11.213} & -- & 18.004 & \best{17.242} & \best{0.055} & \best{2.317} \\
\bottomrule
\end{tabular}}
\end{table*}

\paragraph{Baselines.}
We compare \method{} against ten representative methods under a unified setting. Single-frame baselines are InterWild~\citep{interwild2023}, HaMeR~\citep{hamer2024}, Hamba~\citep{hamba2024}, WiLoR~\citep{wilor2025}, WildHands~\citep{wildhands2024}, and EgoForce~\citep{egoforce2026}. Video-based baselines include OmniHands~\citep{omnihands2025}, Dyn-HaMR~\citep{dynhamr2025}, HaWoR~\citep{hawor2025}, and ViDiHand~\citep{vidihand2026}.

\paragraph{Evaluation protocol.}
All methods are evaluated across sequence-disjoint test splits using identical video segments. Baseline rows in Table~\ref{tab:main} quote the published ViDiHand evaluation under the identical protocol, and EgoForce plus our rows come from the same evaluator. Note that EgoForce$^{\ddagger}$ post-filters its camera-space translation with a causal Kalman filter, and that HaWoR is scored on its camera-space front end with the SLAM and infilling stages disabled (Appendix~\ref{sec:supp_protocol}). HOI4D is held out from our training and evaluated zero-shot, while the H2O and OakInk2 comparisons in Appendix~\ref{sec:supp_results} place our in-domain model against baselines that never saw those datasets.

\paragraph{Metrics.}
Following \citet{vidihand2026}, a hand is defined as on-screen if any of its $21$ ground-truth 3D joints projects within the image boundary at a depth above $z_{\min}=1$\,cm. Frame Accuracy (FAcc) measures the fraction of frames containing neither missed nor spurious hands, while Recall and F1 are calculated per hand. A hand counts as detected when its existence score exceeds $0.5$ and it matches a same-side on-screen ground-truth hand by projected 2D overlap, with the on-screen gate applied to both sides. Because matching is scored under the predicted translation, detection reflects placement as well as presence. MPJPE-p~($\mathrm{mm}$) is the mean joint position error after wrist alignment, and PA-p~($\mathrm{mm}$) the mean joint error following Procrustes alignment. EPE\textsubscript{2D}-p~($\mathrm{px}$) is the mean 2D end-point error over in-frame joints, GO-p~($^{\circ}$) the geodesic global rotation error, and CT-p~($\mathrm{m}$) the camera-frame translation error $\lVert\hat\tau-\tau\rVert_2$. Jitter~($\mathrm{mm}/\mathrm{frame}^2$) is the mean second temporal difference of the 3D joint predictions.

The \mbox{-p} suffix denotes a penalization protocol for false negatives: undetected on-screen hands are assigned a canonical MANO mesh placed at the camera origin, preventing methods from artificially boosting accuracy by dropping challenging detections. Additionally, MPJPE\textsuperscript{+OOS}~($\mathrm{mm}$) evaluates all hands across the sequence including out-of-sight targets, weighted by frame counts, while MPJPE\textsuperscript{OOS}~($\mathrm{mm}$) restricts the same wrist-aligned error to the hand-frames that fail the on-screen gate. The two average over different populations and are not interchangeable. For our $K$-free variant, EPE\textsubscript{2D}-p is evaluated by re-projecting predicted 3D joints using held-out ground-truth intrinsics, directly diagnosing the accuracy of implicit camera calibration, see formal definitions in Appendix~\ref{sec:supp_protocol}.

\subsection{Main Results}
\label{sec:comp}

\paragraph{In-view accuracy.}
\method{} achieves state-of-the-art overall tracking performance across ARCTIC, HOT3D, and HOI4D, leading on $27$ of $29$ evaluation metrics. Detection of the standard configuration is near saturation (FAcc, Recall, and F1 above $0.95$, and $1.00$ on ARCTIC), while seven baselines fall below $0.70$ FAcc on wide-angle HOT3D sequences. Compared with ViDiHand, \method{} reduces MPJPE-p by $30\%$ ($21.67$ to $15.26$\,mm), $40\%$ ($21.51$ to $12.89$\,mm), and $23\%$ ($30.09$ to $23.03$\,mm) on ARCTIC, HOT3D, and zero-shot HOI4D, respectively, and leads on articulation, 2D localization, orientation, and translation throughout. The advantage persists on occlusion-heavy ARCTIC: \method{} reduces PA-p by $24\%$ over ViDiHand and nearly halves the MPJPE-p of OmniHands, a method designed for in-image occlusion recovery. \method{} also yields the lowest Jitter ($2.39$--$3.16\,\mathrm{mm}/\mathrm{frame}^2$), reflecting clip-level decoding and the explicit temporal-smoothness objective.

\paragraph{Intrinsics-free operation.}
The separately trained $K$-free configuration drops test-time calibration entirely while keeping detection at parity with the standard configuration on all five benchmarks, with 2D anchors on par with the calibrated bearings on wide-angle HOT3D (F1 $0.995$ vs.\ $0.996$, EPE\textsubscript{2D}-p $6.07$ vs.\ $6.42$\,px). Its residual cost is a small wrist-aligned pose gap on the narrower-field benchmarks ($+1.36$\,mm MPJPE-p on ARCTIC, $+1.09$ on H2O), while zero-shot HOI4D improves ($22.15$ vs.\ $23.03$\,mm; the higher EPE\textsubscript{2D}-p reflects recovered hard detections entering the matched set).

\begin{figure*}[t]
\centering
\includegraphics[width=\textwidth]{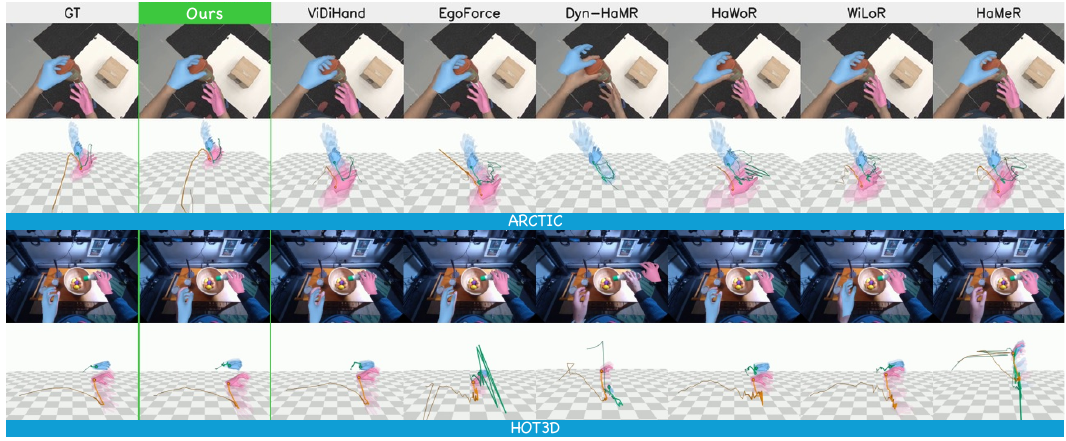}
\vspace{-0.4em}
\caption{\textbf{Qualitative comparison on ARCTIC (top) and HOT3D (bottom).} For each dataset: the predicted mesh on the input frame, and the recovered hands in a shared world frame with wrist trails. Every method predicts in the camera frame, and the world view places those predictions with the ground-truth camera poses, so the view shows placement, not recovered camera motion. Under bimanual occlusion on ARCTIC several baselines recover only one hand. On wide-angle HOT3D they scatter the hands across the floor plane.}
\vspace{-1.0em}
\label{fig:compare}
\end{figure*}

\paragraph{Out-of-sight recovery.}
Once out-of-sight hands enter the evaluation, \method{} achieves $16.8$ and $17.3$\,mm MPJPE\textsuperscript{+OOS} on ARCTIC and HOT3D, reducing the $31.05$ and $44.44$\,mm of ViDiHand by $45.9\%$ and $61.1\%$, respectively. Restricted to the out-of-sight stratum alone, \method{} attains $35.2$ and $38.6$\,mm MPJPE\textsuperscript{OOS} against $149.1$ and $151.7$\,mm for ViDiHand and $135.4$ and $126.4$\,mm for WildHands, the strongest baseline on those frames. The $K$-free variant follows at $18.1$ and $18.7$\,mm MPJPE\textsuperscript{+OOS}, within $1.5$\,mm of the standard configuration. The encoder ablation traces this robustness to the generative VDM features (Table~\ref{tab:vdm}). Note that both metrics are wrist-aligned, reflecting articulation and global orientation through the gap. Absolute out-of-sight placement instead relies on depth and bearing regressed from temporal context through the solver fallback, and remains less constrained (Figure~\ref{fig:compare}).

\paragraph{Efficiency.}
One deterministic pass makes \method{} fast as well as accurate: $63.1$\,fps on one A100 ($63.3$ $K$-free) versus $1.91$\,fps for ViDiHand in its accuracy configuration, a $33\times$ gap (protocol in Appendix~\ref{sec:supp_results}).

\subsection{Ablation Studies}
\label{sec:ablation}

\begin{table}[t]
\centering
\small
\setlength{\tabcolsep}{3pt}
\renewcommand{\arraystretch}{0.79}
\caption{\textbf{Feature-source ablation on ARCTIC.} MPJPE-p, PA-p and MPJPE\textsuperscript{OOS} in mm, Jitter in mm/frame$^2$. MPJPE\textsuperscript{OOS} is restricted to the hand-frames that fail the on-screen gate.}
\label{tab:vdm}
\resizebox{\columnwidth}{!}{%
\begin{tabular}{@{}lcccc@{}}
\toprule
Feature source & MPJPE-p$\downarrow$ & PA-p$\downarrow$ & Jitter$\downarrow$ & MPJPE\textsuperscript{OOS}$\downarrow$ \\
\midrule
\textbf{\method{}} & \best{16.95} & \best{8.16} & \best{2.69} & \best{41.5} \\
\ w/o LoRA                  & 23.10 & 10.83 & 3.35 & 50.1 \\
\ V-JEPA 2~\cite{vjepa2_2025}                  & 17.81 & 8.74 & 4.56 & 66.1 \\
\ VideoMAE~\cite{videomae2022}                  & 22.53 & 10.42 & 5.60 & 77.6 \\
\ VAE latent                & 32.50 & 12.89 & 3.35 & 79.3 \\
\ raw RGB                   & 46.50 & 16.93 & 4.37 & 98.3 \\
\bottomrule
\end{tabular}}
\end{table}

\begin{table}[t]
\centering
\small
\setlength{\tabcolsep}{1.6pt}
\renewcommand{\arraystretch}{0.79}
\caption{\textbf{Camera and decoder ablations on ARCTIC.} MPJPE-p and PA-p in mm, EPE\textsubscript{2D}-p in px, CT-p in m, Jitter in mm/frame$^2$.}
\vspace{-0.3em}
\label{tab:decoder}
\resizebox{\columnwidth}{!}{%
\begin{tabular}{@{}lccccc@{}}
\toprule
Variant & MPJPE-p$\downarrow$ & PA-p$\downarrow$ & EPE\textsubscript{2D}-p$\downarrow$ & CT-p$\downarrow$ & Jitter$\downarrow$ \\
\midrule
\textbf{\method{}} & \best{15.256} & \best{7.474} & \best{9.180} & 0.021 & 2.700 \\
\midrule
\multicolumn{6}{@{}l}{\textit{Removing test-time intrinsics}} \\
\ \method{} ($K$-free) & 15.264 & 7.770 & 13.168 & \best{0.020} & 2.704 \\
\ \ \ w/o mixed-PnP (inverse proj.) & 16.134 & 8.002 & 14.425 & 0.030 & 2.874 \\
\ \ \ w/o PnP solve (direct regr.) & 15.546 & 7.955 & 14.399 & 0.023 & \best{2.658} \\
\midrule
\multicolumn{6}{@{}l}{\textit{Decoder design (ablated from $K$-free)}} \\
\ w/o spatial PE & 15.348 & 8.101 & 13.141 & 0.023 & 2.686 \\
\ w/o joint queries (pooled) & 17.144 & 8.916 & 14.849 & 0.029 & 2.727 \\
\ shape $\hat\beta$ from registers & 15.622 & 7.760 & 14.078 & \best{0.020} & 2.714 \\
\ w/o rotary PE (absolute PE) & 17.787 & 8.671 & 13.915 & 0.033 & 2.948 \\
\bottomrule
\end{tabular}}
\vspace{-0.5em}
\end{table}

\paragraph{Video encoder.}
To assess the necessity of VDM representations in \method{}, we replace VDM features with VAE latents and raw RGB, remove LoRA to test task-specific adaptation, and substitute Wan 2.2 with frozen V-JEPA 2 and VideoMAE encoders to evaluate robustness across feature sources. All variants, including the full model, are trained on ARCTIC only for $10$k steps, a reduced recipe whose MPJPE-p and PA-p sit above the $20$k full-mix numbers of Table~\ref{tab:main}, so the entries are comparable only within this table.

Table~\ref{tab:vdm} demonstrates that \method{} consistently outperforms all alternative feature sources. Compared to raw RGB ($46.50\,\mathrm{mm}$) and VAE latents ($32.50\,\mathrm{mm}$), the full model lowers MPJPE-p to $16.95\,\mathrm{mm}$ and achieves the best MPJPE\textsuperscript{OOS} ($41.5\,\mathrm{mm}$), confirming that high-level spatiotemporal features are essential for temporally coherent recovery. Disabling LoRA adaptation causes MPJPE-p and MPJPE\textsuperscript{OOS} to rise to $23.10\,\mathrm{mm}$ and $50.1\,\mathrm{mm}$, verifying the value of task-specific adaptation. Most importantly, while V-JEPA 2 performs reasonably on visible inputs, its out-of-sight error jumps to $66.1\,\mathrm{mm}$, indicating that generative VDM priors provide superior predictive capabilities when visual cues are missing. Both substitutes run frozen while the reference row adapts Wan 2.2 with LoRA, but the matched frozen-to-frozen contrast points the same way, with V-JEPA 2 at $66.1$ against $50.1\,\mathrm{mm}$ for Wan 2.2 without LoRA.

\paragraph{Decoder and camera solve.}

We ablate the Ray-Based Camera Solver and the Bidirectional Spatiotemporal Decoder on ARCTIC (Table~\ref{tab:decoder}), training all variants under one shared recipe (HOI4D held out). All rows of this table omit $\mathcal{L}_{\mathrm{fit}}$ and read bearings directly from the ray field, so they are comparable within the table but not with Table~\ref{tab:main}. Camera-level ablations evaluate the $K$-free variant, point-wise inverse projection, and direct translation regression. Decoder-level ablations remove spatial PE, replace joint queries with hand-pooled queries, predict clip-level $\hat{\beta}$ via register tokens, or swap rotary relative PE for absolute PE. Two variants, hand-pooled queries and absolute PE, run at half the data-parallel width of the rest at the same step count and the same clips per GPU, so part of their gap may reflect the smaller effective batch rather than the design change alone.

Within this table, the $K$-free row matches the standard configuration ($15.264$ vs.\ $15.256$\,mm MPJPE-p and $0.020$ vs.\ $0.021$\,m CT-p), so learned ray fields replace explicit test-time intrinsics under the camera geometry of ARCTIC. Rotary relative PE and joint queries are pivotal: replacing relative PE with absolute PE degrades MPJPE-p by $2.523\,\mathrm{mm}$ and worsens Jitter from $2.704$ to $2.948\,\mathrm{mm}/\mathrm{frame}^2$, while pooling joint queries adds $1.880\,\mathrm{mm}$ to MPJPE-p and $1.681\,\mathrm{px}$ to EPE\textsubscript{2D}-p. For solver components, replacing mixed-PnP with inverse projection degrades CT-p ($0.020$ to $0.030\,\mathrm{m}$), and direct translation regression increases EPE\textsubscript{2D}-p by $1.231\,\mathrm{px}$. Direct regression does return the lowest Jitter of the table ($2.658\,\mathrm{mm}/\mathrm{frame}^2$), a trade-off we do not adopt, since it loosens image-space placement.

\subsection{Qualitative Comparison}
As illustrated in Figure~\ref{fig:compare}, single-frame and world-space baselines produce fragmented, frequently interrupted predictions, while the generative baseline produces smoother yet depth-biased ones in these examples. In contrast, \method{} recovers continuous, temporally stable trajectories tightly aligned with ground truth under rapid motion, severe occlusion, and transient out-of-sight intervals. Retargeting the recovered trajectories onto a dexterous Inspire Hand yields natural, coordinated motions that closely mirror the source videos, shown as kinematic visualizations in Appendix~\ref{sec:supp_qualapp}, which also extends this comparison to H2O, OakInk2, and two in-the-wild egocentric sources.

\section{Conclusion and Discussion}

We presented \method{}, an offline clip-level framework that recovers metric bimanual hand trajectories from egocentric video. Rather than sampling a video diffusion model as a renderer, we read it once as a Deterministic Clean-Latent Encoder and let 3D supervision reshape its features into geometry-aware ones. A Bidirectional Spatiotemporal Decoder then reconstructs whole clips at once, and a Ray-Based Camera Solver places them metrically. \method{} sets a new state of the art across five benchmarks while running in a single forward pass, $33\times$ faster than the strongest prior method. The best discriminative encoder approaches this accuracy while hands are visible but degrades far more once they leave view, suggesting that generative pretraining may contribute a temporal-geometric prior that discriminative pretraining does not. We hope \method{} serves as a practical bridge from large-scale human video to manipulation data for robot learning.

Several limitations remain. The predicted ray field generalizes best within camera families seen during training, and broader data coverage is the natural remedy. HOI4D and H2O rely partially on pseudo-ground-truth annotations, so higher-precision data is a promising path to further gains. Moreover, \method{} operates offline over complete clips, which suits scalable data curation rather than closed-loop control. Distilling the encoder into a streaming variant is a natural step toward online deployment. 

{
    \small
    \setlength{\bibsep}{0.5pt} % tighten reference spacing so the list ends on this page
    \bibliographystyle{ieeenat_fullname}
    \bibliography{main}

@String(CVPR= {IEEE Conf. Comput. Vis. Pattern Recog.})

@String(ICCV= {Int. Conf. Comput. Vis.})

@String(ECCV= {Eur. Conf. Comput. Vis.})

@String(ICLR = {Int. Conf. Learn. Represent.})

@String(CVPR  = {CVPR})

@String(ICCV  = {ICCV})

@String(ECCV  = {ECCV})

@String(ICLR  = {ICLR})

@misc{vidihand2026,
  author        = {Wang, Yuxi and Jin, Chengkai and Liu, Yufei and Ouyang, Wenqi and Wei, Tianyi and Zeng, Zhiwei and Huang, Siyuan and Shen, Zhiqi and Pan, Xingang},
  title         = {The Surprising Effectiveness of Video Diffusion Models for Hand Motion Reconstruction},
  year          = {2026},
  eprint        = {2606.30308},
  archiveprefix = {arXiv},
  primaryclass  = {cs.CV}
}

@inproceedings{hamer2024,
  author    = {Pavlakos, Georgios and Shan, Dandan and Radosavovic, Ilija and Kanazawa, Angjoo and Fouhey, David and Malik, Jitendra},
  title     = {Reconstructing Hands in {3D} with Transformers},
  booktitle = {Proceedings of the IEEE/CVF Conference on Computer Vision and Pattern Recognition (CVPR)},
  year      = {2024}
}

@inproceedings{wilor2025,
  author    = {Potamias, Rolandos Alexandros and Zhang, Jinglei and Deng, Jiankang and Zafeiriou, Stefanos},
  title     = {{WiLoR}: End-to-end {3D} Hand Localization and Reconstruction in-the-wild},
  booktitle = {Proceedings of the IEEE/CVF Conference on Computer Vision and Pattern Recognition (CVPR)},
  year      = {2025},
  note      = {arXiv:2409.12259}
}

@inproceedings{interwild2023,
  author    = {Moon, Gyeongsik},
  title     = {Bringing Inputs to Shared Domains for {3D} Interacting Hands Recovery in the Wild},
  booktitle = {Proceedings of the IEEE/CVF Conference on Computer Vision and Pattern Recognition (CVPR)},
  year      = {2023},
  note      = {arXiv:2303.13652}
}

@inproceedings{wildhands2024,
  author    = {Prakash, Aditya and Tu, Ruisen and Chang, Matthew and Gupta, Saurabh},
  title     = {{3D} Hand Pose Estimation in Everyday Egocentric Images},
  booktitle = {Proceedings of the European Conference on Computer Vision (ECCV)},
  year      = {2024}
}

@inproceedings{hamba2024,
  author    = {Dong, Haoye and Chharia, Aviral and Gou, Wenbo and Vicente Carrasco, Francisco and De la Torre, Fernando},
  title     = {Hamba: Single-view {3D} Hand Reconstruction with Graph-guided Bi-Scanning Mamba},
  booktitle = {Advances in Neural Information Processing Systems (NeurIPS)},
  year      = {2024}
}

@misc{omnihands2025,
  author        = {Lin, Dixuan and Zhang, Yuxiang and Li, Mengcheng and Jing, Wei and Yan, Qi and Wang, Qianying and Liu, Yebin and Zhang, Hongwen},
  title         = {{OmniHands}: Towards Robust {4D} Hand Mesh Recovery via A Versatile Transformer},
  year          = {2024},
  eprint        = {2405.20330},
  archiveprefix = {arXiv},
  primaryclass  = {cs.CV}
}

@inproceedings{dynhamr2025,
  author    = {Yu, Zhengdi and Zafeiriou, Stefanos and Birdal, Tolga},
  title     = {{Dyn-HaMR}: Recovering {4D} Interacting Hand Motion from a Dynamic Camera},
  booktitle = {Proceedings of the IEEE/CVF Conference on Computer Vision and Pattern Recognition (CVPR)},
  year      = {2025},
  note      = {arXiv:2412.12861}
}

@inproceedings{hawor2025,
  author    = {Zhang, Jinglei and Deng, Jiankang and Ma, Chao and Potamias, Rolandos Alexandros},
  title     = {{HaWoR}: World-Space Hand Motion Reconstruction from Egocentric Videos},
  booktitle = {Proceedings of the IEEE/CVF Conference on Computer Vision and Pattern Recognition (CVPR)},
  year      = {2025},
  note      = {arXiv:2501.02973}
}

@inproceedings{egoforce2026,
  author    = {Millerdurai, Christen and Wang, Shaoxiang and Xie, Yaxu and Golyanik, Vladislav and Stricker, Didier and Pagani, Alain},
  title     = {{EgoForce}: Forearm-Guided Camera-Space {3D} Hand Pose from a Monocular Egocentric Camera},
  booktitle = {ACM SIGGRAPH 2026 Conference Papers},
  year      = {2026},
  note      = {arXiv:2605.12498}
}

@inproceedings{tram2024,
  author    = {Wang, Yufu and Wang, Ziyun and Liu, Lingjie and Daniilidis, Kostas},
  title     = {{TRAM}: Global Trajectory and Motion of {3D} Humans from In-the-wild Videos},
  booktitle = {Proceedings of the European Conference on Computer Vision (ECCV)},
  year      = {2024},
  note      = {arXiv:2403.17346}
}

@inproceedings{genmo2025,
  author    = {Li, Jiefeng and Cao, Jinkun and Zhang, Haotian and Rempe, Davis and Kautz, Jan and Iqbal, Umar and Yuan, Ye},
  title     = {{GENMO}: A Generalist Model for Human Motion},
  booktitle = {Proceedings of the IEEE/CVF International Conference on Computer Vision (ICCV)},
  year      = {2025},
  note      = {arXiv:2505.01425}
}

@misc{genception2026,
  author        = {Wang, Letian and Zhang, Chuhan and Kabra, Rishabh and Uijlings, Jasper and Waslander, Steven and Zisserman, Andrew and Carreira, Joao and He, Kaiming and Andriluka, Misha and Bazavan, Eduard Gabriel and Zanfir, Andrei and Sminchisescu, Cristian},
  title         = {Video Generation Models are General-Purpose Vision Learners},
  year          = {2026},
  eprint        = {2607.09024},
  archiveprefix = {arXiv},
  primaryclass  = {cs.CV},
  note          = {GenCeption; Google DeepMind}
}

@inproceedings{egomimic2024,
  author    = {Kareer, Simar and Patel, Dhruv and Punamiya, Ryan and Mathur, Pranay and Cheng, Shuo and Wang, Chen and Hoffman, Judy and Xu, Danfei},
  title     = {{EgoMimic}: Scaling Imitation Learning via Egocentric Video},
  booktitle = {IEEE International Conference on Robotics and Automation (ICRA)},
  year      = {2025},
  note      = {arXiv:2410.24221}
}

@misc{egovla2025,
  author        = {Yang, Ruihan and Yu, Qinxi and Wu, Yecheng and Yan, Rui and Li, Borui and Cheng, An-Chieh and Zou, Xueyan and Fang, Yunhao and Cheng, Xuxin and Qiu, Ri-Zhao and Yin, Hongxu and Liu, Sifei and Han, Song and Lu, Yao and Wang, Xiaolong},
  title         = {{EgoVLA}: Learning Vision-Language-Action Models from Egocentric Human Videos},
  year          = {2025},
  eprint        = {2507.12440},
  archiveprefix = {arXiv},
  primaryclass  = {cs.RO}
}

@article{mano2017,
  author  = {Romero, Javier and Tzionas, Dimitrios and Black, Michael J.},
  title   = {Embodied Hands: Modeling and Capturing Hands and Bodies Together},
  journal = {ACM Transactions on Graphics},
  volume  = {36},
  number  = {6},
  year    = {2017},
  note    = {Proc. SIGGRAPH Asia 2017; the MANO hand model}
}

@inproceedings{arctic2023,
  author    = {Fan, Zicong and Taheri, Omid and Tzionas, Dimitrios and Kocabas, Muhammed and Kaufmann, Manuel and Black, Michael J. and Hilliges, Otmar},
  title     = {{ARCTIC}: A Dataset for Dexterous Bimanual Hand-Object Manipulation},
  booktitle = {Proceedings of the IEEE/CVF Conference on Computer Vision and Pattern Recognition (CVPR)},
  year      = {2023}
}

@inproceedings{hot3d2024,
  author    = {Banerjee, Prithviraj and Shkodrani, Sindi and Moulon, Pierre and Hampali, Shreyas and Han, Shangchen and Zhang, Fan and Zhang, Linguang and Fountain, Jade and Miller, Edward and Basol, Selen and Newcombe, Richard and Wang, Robert and Engel, Jakob Julian and Hodan, Tomas},
  title     = {{HOT3D}: Hand and Object Tracking in {3D} from Egocentric Multi-View Videos},
  booktitle = {Proceedings of the IEEE/CVF Conference on Computer Vision and Pattern Recognition (CVPR)},
  year      = {2025},
  note      = {arXiv:2411.19167}
}

@inproceedings{hoi4d2022,
  author    = {Liu, Yunze and Liu, Yun and Jiang, Che and Lyu, Kangbo and Wan, Weikang and Shen, Hao and Liang, Boqiang and Fu, Zhoujie and Wang, He and Yi, Li},
  title     = {{HOI4D}: A {4D} Egocentric Dataset for Category-Level Human-Object Interaction},
  booktitle = {Proceedings of the IEEE/CVF Conference on Computer Vision and Pattern Recognition (CVPR)},
  year      = {2022}
}

@inproceedings{h2o2021,
  author    = {Kwon, Taein and Tekin, Bugra and St{\"u}hmer, Jan and Bogo, Federica and Pollefeys, Marc},
  title     = {{H2O}: Two Hands Manipulating Objects for First Person Interaction Recognition},
  booktitle = {Proceedings of the IEEE/CVF International Conference on Computer Vision (ICCV)},
  year      = {2021}
}

@inproceedings{oakink2_2024,
  author    = {Zhan, Xinyu and Yang, Lixin and Zhao, Yifei and Mao, Kangrui and Xu, Hanlin and Lin, Zenan and Li, Kailin and Lu, Cewu},
  title     = {{OakInk2}: A Dataset of Bimanual Hands-Object Manipulation in Complex Task Completion},
  booktitle = {Proceedings of the IEEE/CVF Conference on Computer Vision and Pattern Recognition (CVPR)},
  year      = {2024}
}

@inproceedings{egodex2025,
  author    = {Hoque, Ryan and Huang, Peide and Yoon, David J. and Sivapurapu, Mouli and Zhang, Jian},
  title     = {{EgoDex}: Learning Dexterous Manipulation from Large-Scale Egocentric Video},
  booktitle = {International Conference on Learning Representations (ICLR)},
  year      = {2026},
  note      = {arXiv:2505.11709. Apple}
}

@inproceedings{freihand2019,
  author    = {Zimmermann, Christian and Ceylan, Duygu and Yang, Jimei and Russell, Bryan and Argus, Max and Brox, Thomas},
  title     = {{FreiHAND}: A Dataset for Markerless Capture of Hand Pose and Shape from Single {RGB} Images},
  booktitle = {Proceedings of the IEEE/CVF International Conference on Computer Vision (ICCV)},
  year      = {2019}
}

@inproceedings{rhd2017,
  author    = {Zimmermann, Christian and Brox, Thomas},
  title     = {Learning to Estimate {3D} Hand Pose from Single {RGB} Images},
  booktitle = {Proceedings of the IEEE International Conference on Computer Vision (ICCV)},
  year      = {2017}
}

@inproceedings{reinterhand2023,
  author    = {Moon, Gyeongsik and Saito, Shunsuke and Xu, Weipeng and others},
  title     = {A Dataset of Relighted {3D} Interacting Hands},
  booktitle = {Advances in Neural Information Processing Systems (NeurIPS) Datasets and Benchmarks},
  year      = {2023}
}

@misc{wan2025,
  author        = {{Wan Team}},
  title         = {Wan: Open and Advanced Large-Scale Video Generative Models},
  year          = {2025},
  eprint        = {2503.20314},
  archiveprefix = {arXiv},
  primaryclass  = {cs.CV},
  note          = {Alibaba; covers the Wan2.1/Wan2.2 model family}
}

@inproceedings{vace2025,
  author    = {Jiang, Zeyinzi and Han, Zhen and Mao, Chaojie and Zhang, Jingfeng and Pan, Yulin and Liu, Yu},
  title     = {{VACE}: All-in-One Video Creation and Editing},
  booktitle = {Proceedings of the IEEE/CVF International Conference on Computer Vision (ICCV)},
  year      = {2025},
  note      = {arXiv:2503.07598}
}

@inproceedings{lora2022,
  author    = {Hu, Edward J. and Shen, Yelong and Wallis, Phillip and Allen-Zhu, Zeyuan and Li, Yuanzhi and Wang, Shean and Wang, Lu and Chen, Weizhu},
  title     = {{LoRA}: Low-Rank Adaptation of Large Language Models},
  booktitle = {International Conference on Learning Representations (ICLR)},
  year      = {2022}
}

@misc{vjepa2_2025,
  author        = {Assran, Mido and Bardes, Adrien and Fan, David and Garrido, Quentin and Howes, Russell and Komeili, Mojtaba and others},
  title         = {{V-JEPA} 2: Self-Supervised Video Models Enable Understanding, Prediction and Planning},
  year          = {2025},
  eprint        = {2506.09985},
  archiveprefix = {arXiv},
  primaryclass  = {cs.CV},
  note          = {Meta AI}
}

@inproceedings{detr2020,
  author    = {Carion, Nicolas and Massa, Francisco and Synnaeve, Gabriel and Usunier, Nicolas and Kirillov, Alexander and Zagoruyko, Sergey},
  title     = {End-to-End Object Detection with Transformers},
  booktitle = {Proceedings of the European Conference on Computer Vision (ECCV)},
  year      = {2020}
}

@article{kuhn1955,
  author  = {Kuhn, Harold W.},
  title   = {The {H}ungarian Method for the Assignment Problem},
  journal = {Naval Research Logistics Quarterly},
  volume  = {2},
  year    = {1955}
}

@inproceedings{pointprompting2026,
  author    = {Shrivastava, Ayush and Mehta, Sanyam and Geng, Daniel and Owens, Andrew},
  title     = {Point Prompting: Counterfactual Tracking with Video Diffusion Models},
  booktitle = {International Conference on Learning Representations (ICLR)},
  year      = {2026},
  note      = {arXiv:2510.11715}
}

@article{rope2024,
  author  = {Su, Jianlin and Ahmed, Murtadha and Lu, Yu and Pan, Shengfeng and Bo, Wen and Liu, Yunfeng},
  title   = {{RoFormer}: Enhanced Transformer with Rotary Position Embedding},
  journal = {Neurocomputing},
  volume  = {568},
  year    = {2024}
}

@inproceedings{vibe2020,
  author    = {Kocabas, Muhammed and Athanasiou, Nikos and Black, Michael J.},
  title     = {{VIBE}: Video Inference for Human Body Pose and Shape Estimation},
  booktitle = {Proceedings of the IEEE/CVF Conference on Computer Vision and Pattern Recognition (CVPR)},
  year      = {2020}
}

@inproceedings{tcmr2021,
  author    = {Choi, Hongsuk and Moon, Gyeongsik and Chang, Ju Yong and Lee, Kyoung Mu},
  title     = {Beyond Static Features for Temporally Consistent {3D} Human Pose and Shape from a Video},
  booktitle = {Proceedings of the IEEE/CVF Conference on Computer Vision and Pattern Recognition (CVPR)},
  year      = {2021}
}

@inproceedings{glot2023,
  author    = {Shen, Xiaolong and Yang, Zongxin and Wang, Xiaohan and Ma, Jianxin and Zhou, Chang and Yang, Yi},
  title     = {Global-to-Local Modeling for Video-based {3D} Human Pose and Shape Estimation},
  booktitle = {Proceedings of the IEEE/CVF Conference on Computer Vision and Pattern Recognition (CVPR)},
  year      = {2023}
}

@inproceedings{slahmr2023,
  author    = {Ye, Vickie and Pavlakos, Georgios and Malik, Jitendra and Kanazawa, Angjoo},
  title     = {Decoupling Human and Camera Motion from Videos in the Wild},
  booktitle = {Proceedings of the IEEE/CVF Conference on Computer Vision and Pattern Recognition (CVPR)},
  year      = {2023}
}

@inproceedings{mdm2023,
  author    = {Tevet, Guy and Raab, Sigal and Gordon, Brian and Shafir, Yoni and Cohen-Or, Daniel and Bermano, Amit H.},
  title     = {Human Motion Diffusion Model},
  booktitle = {International Conference on Learning Representations (ICLR)},
  year      = {2023}
}

@misc{egoh4_2025,
  author        = {Hatano, Masashi and Zhu, Zhifan and Saito, Hideo and Damen, Dima},
  title         = {The Invisible {EgoHand}: {3D} Hand Forecasting through {EgoBody} Pose Estimation},
  year          = {2025},
  eprint        = {2504.08654},
  archiveprefix = {arXiv},
  primaryclass  = {cs.CV}
}

@inproceedings{videomae2022,
  author    = {Tong, Zhan and Song, Yibing and Wang, Jue and Wang, Limin},
  title     = {{VideoMAE}: Masked Autoencoders are Data-Efficient Learners for Self-Supervised Video Pre-Training},
  booktitle = {Advances in Neural Information Processing Systems (NeurIPS)},
  year      = {2022}
}

@inproceedings{raydiffusion2024,
  author    = {Zhang, Jason Y. and Lin, Amy and Kumar, Moneish and Yang, Tzu-Hsuan and Ramanan, Deva and Tulsiani, Shubham},
  title     = {Cameras as Rays: Pose Estimation via Ray Diffusion},
  booktitle = {International Conference on Learning Representations (ICLR)},
  year      = {2024}
}

@inproceedings{perspectivefields2023,
  author    = {Jin, Linyi and Zhang, Jianming and Hold-Geoffroy, Yannick and Wang, Oliver and Matzen, Kevin and Sticha, Matthew and Fouhey, David F.},
  title     = {Perspective Fields for Single Image Camera Calibration},
  booktitle = {Proceedings of the IEEE/CVF Conference on Computer Vision and Pattern Recognition (CVPR)},
  year      = {2023}
}

@inproceedings{holoassist2023,
  author    = {Wang, Xin and Kwon, Taein and Rad, Mahdi and Pan, Bowen and Chakraborty, Ishani and Andrist, Sean and Bohus, Dan and Feniello, Ashley and Tekin, Bugra and Frujeri, Felipe Vieira and Joshi, Neel and Pollefeys, Marc},
  title     = {{HoloAssist}: An Egocentric Human Interaction Dataset for Interactive {AI} Assistants in the Real World},
  booktitle = {Proceedings of the IEEE/CVF International Conference on Computer Vision (ICCV)},
  year      = {2023}
}

@misc{xperience10m2026,
  author       = {{Ropedia}},
  title        = {{Xperience-10M}: A Large-Scale Egocentric Multimodal Dataset with Structured 3D/4D Annotations},
  howpublished = {\url{https://huggingface.co/datasets/ropedia-ai/xperience-10m}},
  year         = {2026},
  note         = {Dataset}
}

@misc{li2026aceego0unifyingegocentrichuman,
  author        = {Li, Hao and Zhao, Ganlong and Liu, Yufei and Hou, Haotian and Ye, Guoquan and Fang, Tongyan and Liu, Chunxiao and Huang, Siyuan and Liu, Jianbo and Wang, Xiaogang and Li, Hongsheng},
  title         = {{ACE-Ego-0}: Unifying Egocentric Human and Robotic Data for {VLA} Pretraining},
  year          = {2026},
  eprint        = {2606.17200},
  archiveprefix = {arXiv},
  primaryclass  = {cs.RO}
}
}

\appendix
% Appendix floats and displays use the ``S'' series so they cannot be confused with
% the main text's Tables 1-3, Figures 1-4 and Equations 1-6.
\renewcommand{\thetable}{S\arabic{table}}\setcounter{table}{0}
\renewcommand{\thefigure}{S\arabic{figure}}\setcounter{figure}{0}
\renewcommand{\theequation}{S\arabic{equation}}\setcounter{equation}{0}
% Insurance: keep hyperref's anchor names in step with the S-series. With the caption
% package loaded (cvpr.sty) anchors are caption-based and this is a no-op, but it costs
% nothing and prevents a table.1/figure.1 collision if that ever changes.
\def\theHtable{S\arabic{table}}
\def\theHfigure{S\arabic{figure}}
\def\theHequation{S\arabic{equation}}
% cvpr.sty provides this: a full-width title plus ``Supplementary Material'' header on a new page.
\maketitlesupplementary

The appendix is organized as follows. Appendix~\ref{sec:supp_protocol} states the evaluation protocol, the dataset splits, and how the baselines were reproduced, so that every number in the main text can be traced to a definition. Appendix~\ref{sec:supp_impl} gives the architecture, optimization, and loss details needed to reproduce the model. Appendix~\ref{sec:supp_results} reports the two additional in-domain datasets, the out-of-sight stratification behind the two OOS metrics of the main text, the matched-detection check, and inference throughput. Appendix~\ref{sec:supp_analysis} analyzes the image-periphery failure that motivates the $K$-free camera fit and the cost of long decoding passes, and measures two design choices, the tap depth and the latent noise level. Appendix~\ref{sec:supp_qualapp} gives extended qualitative comparisons and the retargeting application. Appendix~\ref{sec:supp_limits} closes with the limitations, the failure signature each one produces, and our data-use statement.

\section{Experimental Setup and Protocol}
\label{sec:supp_protocol}
\subsection{Metric Definitions and Penalty Protocol}
This section states the definitions behind every result table of the main text. Let a hand be a pair $(J,\tau)$ of $21$ canonical joints and a camera-frame translation, and write $J^{\mathrm{cam}}=J+\tau$. Note that $\tau$ places the canonical hand and is therefore the MANO root translation, not the position of the wrist joint, which is $J_0+\tau$. CT below is an error in $\tau$.

Table~\ref{tab:metrics} lists every metric symbol used in this paper, the population it averages over, and whether the coverage penalty applies. The rest of this section defines each one. One further metric appears in Appendix~\ref{sec:supp_results} and is named in place: MPJPE restricted to \emph{matched} detections, a third population again, used there to separate lost coverage from lost pose.

\paragraph{On-screen gate.} A hand is \emph{on-screen} if at least one of its $21$ joints projects inside the image rectangle at positive depth,
\begin{equation}
\exists j:\; \pi(J^{\mathrm{cam}}_j)\in[0,W)\times[0,H) \;\wedge\; J^{\mathrm{cam}}_{z,j}>z_{\min},
\end{equation}
with $\pi$ the calibrated projection and $z_{\min}=1$\,cm. Ground-truth hands that fail this test are out of sight (OOS) and are excluded from the main metrics. The same test is applied to predictions, so a prediction that places its hand OOS is not charged as a false positive.

\paragraph{Detection.} A predicted hand is active when its existence score exceeds $0.5$. Active predictions are matched to on-screen ground-truth hands of the same side by the intersection-over-union of their projected mesh bounding boxes, with ground-truth boxes dilated by $10\%$ before the overlap is computed. Each prediction takes its highest-overlap candidate, and where two same-side predictions select the same ground-truth hand, the higher overlap wins. A prediction whose best overlap is zero, or whose best overlap falls on the opposite side, is counted as a false positive, so the rule is a strictly-positive-overlap test on the matching side rather than a fixed IoU threshold. Unmatched ground-truth hands count as false negatives (FN), and unmatched predictions as false positives (FP). With per-side counts, and writing $\mathrm{P}$ and $\mathrm{R}$ for precision and recall,
\begin{equation}
\mathrm{P}=\frac{\mathrm{TP}}{\mathrm{TP}+\mathrm{FP}},\qquad
\mathrm{R}=\frac{\mathrm{TP}}{\mathrm{TP}+\mathrm{FN}},\qquad
\mathrm{F1}=\frac{2\,\mathrm{P}\,\mathrm{R}}{\mathrm{P}+\mathrm{R}},
\end{equation}
and Frame Accuracy is the fraction of frames that contain no error on either side,
\begin{equation}
\mathrm{FAcc}=\frac{1}{N}\sum_{f=1}^{N}\mathbf{1}\!\left[\mathrm{FP}_f+\mathrm{FN}_f=0\right].
\end{equation}
Because matching uses the projected mesh, detection depends on predicted translation as well as on the existence score.

\paragraph{Pose metrics.} For a matched pair, with $\bar J=J-J_{0}$ denoting wrist-relative joints and $\Lambda$ the Procrustes alignment,
\begin{align}
\mathrm{MPJPE}&=\tfrac{1}{21}\textstyle\sum_j \lVert \bar{\hat J}_j-\bar J_j\rVert_2,\\
\mathrm{PA}&=\tfrac{1}{21}\textstyle\sum_j \lVert \Lambda(\hat J)_j-J_j\rVert_2,\\
\mathrm{EPE_{2D}}&=\tfrac{1}{|\mathcal{V}|}\textstyle\sum_{j\in\mathcal{V}} \lVert \hat p_j-p_j\rVert_2,\\
\mathrm{CT}&=\lVert \hat\tau-\tau\rVert_2,
\end{align}
where $\hat p_j$ is the model's own 2D anchor, $\mathcal{V}$ holds the joints whose ground truth falls inside the frame, and $\mathrm{GO}=\angle(\hat R,R)$ is the geodesic angle between predicted and ground-truth global orientations. One case needs stating separately. For the $K$-free configuration no intrinsics are supplied to fix a pixel scale, so its $\hat p_j$ is obtained by re-projecting the predicted camera-frame joints under the clip's ground-truth intrinsics. Those intrinsics are used for scoring only and never reach the model, so that EPE\textsubscript{2D}-p becomes a direct read-out of how well the predicted ray field stands in for calibration. Jitter is the mean second temporal difference of the assembled joints $\tilde J=\bar{\hat J}+\hat\tau$, the wrist-relative joint set plus the predicted translation, $\tfrac{1}{T-2}\sum_t\lVert \tilde J_{t+1}-2\tilde J_{t}+\tilde J_{t-1}\rVert$, in $\mathrm{mm}/\mathrm{frame}^2$. The difference runs over the frame index and is not normalized by frame rate. Jitter is evaluated only on frames for which a method produced a matched detection. For each segment and each ground-truth hand the matched predictions are partitioned into maximal runs of consecutive frames, and the second difference is averaged inside every run of at least three frames. A frame with no detection ends a run rather than being filled by interpolation or by a placeholder pose, and no run crosses a segment boundary. A missed detection therefore removes a frame from the computation instead of charging it, so the convention is conservative for a method that detects nearly every hand.

\paragraph{Reading Jitter.} The Bidirectional Spatiotemporal Decoder emits one token set per latent frame, and the Wan VAE compresses time by $4\times$, so the $81$ frames scored in Table~\ref{tab:main} are carried by $21$ latent frames. Hidden features and the direct joint coordinates are expanded to frame rate by linear interpolation along time, and the per-frame Pose Head and Camera Head then run on the expanded features, while the Shape Head reads a single $\hat\beta$ per hand per clip. The scored joints are therefore decoded per frame from a feature trajectory that is piecewise linear in time. Jitter, in $\mathrm{mm}/\mathrm{frame}^2$, measures the smoothness of that output. Fidelity to the ground truth is judged by the accuracy columns reported beside Jitter.

\paragraph{Coverage penalty (the \mbox{-p} suffix).} Reporting errors only on matched hands rewards a method for skipping the hardest frames. Every metric therefore averages over true positives \emph{and} false negatives, charging each FN the error of a canonical MANO~\citep{mano2017} hand: identity global orientation, zero articulation, mean shape, and $\tau=\mathbf{0}$,
\begin{equation}
\mathrm{metric}\text{-}p=\frac{\sum_{i\in\mathrm{TP}} e_i+\sum_{i\in\mathrm{FN}} e^{\mathrm{can}}_i}{|\mathrm{TP}|+|\mathrm{FN}|}.
\end{equation}
The placeholder is computed per missed ground-truth hand, so the cost of a miss is deterministic and identical for every method. $\mathrm{EPE_{2D}}$ is the one exception. Each joint of a missed hand is charged the image diagonal $\sqrt{W^2+H^2}$, an upper bound on the distance between two points inside the frame. That diagonal is $826$\,px on ARCTIC and $679$\,px on HOT3D.

\begin{table}[t]
\centering
\small
\setlength{\tabcolsep}{2.4pt}
\renewcommand{\arraystretch}{0.9}
\caption{\textbf{Metric symbols.} ``Averaged over'' is the population each metric runs on. ``Pen.'' marks the metrics that charge a placeholder for every missed on-screen detection. The upper block is detection-gated: restricted to on-screen ground truth, matched to predictions, with the \mbox{-p} metrics penalized for each miss. The lower block is ground-truth-gated: no matching, no placeholder, and the two differ only in which hand-frames they average over. The two families are not interchangeable.}
\label{tab:metrics}
\begin{tabular}{@{}l l c l@{}}
\toprule
Symbol & Averaged over & Pen. & Unit \\
\midrule
FAcc & frames & -- & -- \\
Recall, F1 & on-screen hands & -- & -- \\
MPJPE-p & on-screen hands & \checkmark & mm \\
PA-p & on-screen hands & \checkmark & mm \\
EPE\textsubscript{2D}-p & on-screen joints & \checkmark & px \\
GO-p & on-screen hands & \checkmark & $^{\circ}$ \\
CT-p & on-screen hands & \checkmark & m \\
Jitter & matched runs & -- & mm/frame$^2$ \\
\midrule
MPJPE\textsuperscript{OOS} & out-of-sight hand-frames & -- & mm \\
MPJPE\textsuperscript{+OOS} & all hand-frames & -- & mm \\
\bottomrule
\end{tabular}
\end{table}

\paragraph{Coverage of out-of-sight hands: two distinct metrics.} The on-screen gate restricts every metric above to ground-truth hands inside the frame, so none of them scores a hand while it is out of sight. Two further metrics close that gap. They are easy to confuse, so we state what separates them before defining either.

Both come from a second scoring pass that is gated by the ground truth alone. That pass walks every ground-truth hand-frame of the sequence and scores it against whatever the model emits in that hand slot, with no existence gate, no detection matching, and no false-negative placeholder. Neither metric therefore carries the \mbox{-p} coverage penalty, and neither is a function of MPJPE-p: MPJPE-p averages over matched detections plus a placeholder for each miss, whereas the two out-of-sight metrics average over ground-truth hand-frames regardless of whether the method detected anything, so the two scoring passes cannot be recombined arithmetically. Both are wrist-aligned, so they certify articulation and orientation through an out-of-sight interval rather than absolute placement.

Write $\mathcal{H}$ for the ground-truth hand-frames of a sequence, $\mathcal{H}^{\mathrm{IV}}\subseteq\mathcal{H}$ for those that pass the on-screen gate, the in-view stratum, and $\mathcal{H}^{\mathrm{OOS}}=\mathcal{H}\setminus\mathcal{H}^{\mathrm{IV}}$ for those that fail it, and let
\begin{equation}
\varepsilon_i=\frac{1}{21}\sum_j \lVert \bar{\hat J}^{(i)}_j-\bar J^{(i)}_j\rVert_2
\end{equation}
be the wrist-aligned error of hand-frame $i$, written $\varepsilon$ to keep it distinct from the generic per-hand error $e$ of the penalty above. The two metrics are the same error averaged over different populations,
\begin{equation}
\begin{aligned}
\mathrm{MPJPE}^{\mathrm{OOS}}&=\frac{1}{|\mathcal{H}^{\mathrm{OOS}}|}\sum_{i\in\mathcal{H}^{\mathrm{OOS}}} \varepsilon_i,\\
\mathrm{MPJPE}^{+\mathrm{OOS}}&=\frac{1}{|\mathcal{H}|}\sum_{i\in\mathcal{H}} \varepsilon_i.
\end{aligned}
\end{equation}
The second is therefore the frame-count-weighted mixture of the two strata. Writing $\bar\varepsilon_{\mathrm{IV}}$ and $\bar\varepsilon_{\mathrm{OOS}}$ for the mean of $\varepsilon_i$ over each stratum, and $n_{\mathrm{IV}}$ and $n_{\mathrm{OOS}}$ for the two hand-frame counts,
\begin{equation}
\mathrm{MPJPE}^{+\mathrm{OOS}}=\frac{n_{\mathrm{IV}}\,\bar\varepsilon_{\mathrm{IV}}+n_{\mathrm{OOS}}\,\bar\varepsilon_{\mathrm{OOS}}}{n_{\mathrm{IV}}+n_{\mathrm{OOS}}}.
\label{eq:oosmix}
\end{equation}
MPJPE\textsuperscript{OOS} scores the out-of-sight stratum on its own. MPJPE\textsuperscript{+OOS} pools both strata, so it is dominated by the larger one: out-of-sight hand-frames are $6.9\%$ of ARCTIC, $17.6\%$ of HOT3D, and $14.5\%$ of OakInk2, so the blended figure is dominated by the in-view stratum on all three. A reduction on MPJPE\textsuperscript{+OOS} is therefore \emph{not} a reduction on out-of-sight hands, and the two must not be read as substitutes for one another. Table~\ref{tab:main} (the main comparison) reports the blended form, Table~\ref{tab:vdm} (the feature-source ablation) reports the stratum, and Table~\ref{tab:oos} of this appendix reports both, together with the in-view stratum, so that Eq.~\eqref{eq:oosmix} can be checked row by row.

\subsection{Datasets, Splits, and Preprocessing}
The five evaluated video datasets (ARCTIC~\citep{arctic2023}, HOT3D~\citep{hot3d2024}, H2O~\citep{h2o2021}, OakInk2~\citep{oakink2_2024}, and HOI4D~\citep{hoi4d2022}) are processed at $30$\,fps without temporal subsampling. On the four of these that enter training, training operates on full-length recordings, drawing a random $21$-latent-frame ($81$ RGB frame) window from each recording at every step, while evaluation decodes the fixed $81$-frame test segments shared with all baselines. Three further datasets enter the training mixture only. They do not all follow this convention and are described at the end of this subsection. Input resolutions are $672\times480$ for ARCTIC and $480\times480$ for HOT3D, and H2O and OakInk2 are resized to a width of $832$ with intrinsics rescaled accordingly, so that every dataset yields an even latent grid. Test splits are subject-disjoint on ARCTIC (test subject s05) and H2O (test subject 4), recording-level on HOT3D ($126$/$72$ recordings), and sequence-level on OakInk2 (evaluated on the $202$-segment subset shared with the baselines), while HOI4D is excluded from training entirely and evaluated zero-shot. Each dataset ships its own hand annotation format, and we convert all of them into one shared MANO format so that a single loader and a single evaluator serve every dataset. As the main text notes, HOI4D and H2O rely partially on pseudo-ground-truth MANO annotation derived from a per-frame estimator, so on those two benchmarks the comparison measures agreement with the labels rather than absolute accuracy. The two differ in how much that matters. HOI4D is held out of training entirely, so its labels are noisy but the noise is the same for every method. H2O is in our training mixture, so our model additionally trains on the label distribution that later scores it, while every baseline meets that distribution for the first time at evaluation. ARCTIC, HOT3D, and OakInk2 carry native MANO ground truth, and the headline reductions are quoted on ARCTIC and HOT3D.

\paragraph{Training mixture.} Both configurations train on the same seven-source mixture, with a dataset drawn per batch from the weights of Table~\ref{tab:mix}, which sum to $100$. Four sources are egocentric video with MANO, namely ARCTIC, HOT3D, H2O, and OakInk2. The remaining three are used for training only and are never evaluated on. They hold $44\%$ of the sampling weight and add appearance and pose diversity from outside the egocentric video domain. The three differ in kind. Re:InterHand~\citep{reinterhand2023} contributes relit studio captures rendered under egocentric fisheye cameras at $10$\,fps, so it enters as video and receives the full $21$-latent-frame window like the video datasets above. FreiHAND~\citep{freihand2019} and RHD~\citep{rhd2017} are two static image sources. Each image is replicated into a static five-frame clip. Those batches contain almost no temporal variation, so they supply appearance and hand-pose diversity rather than motion. FreiHAND is right-hand only and carries a full MANO fit, as does Re:InterHand. RHD provides $21$ 3D joints and no MANO fit. Those samples supervise the 2D anchor, 3D joint, existence, and visibility heads while the rotation and shape terms are held at zero, following the loss routing described in the main text. Every batch is drawn from a single dataset, so clips of different length and different latent grid are never stacked together.

\begin{table}[!t]
\centering
\small
\setlength{\tabcolsep}{1.6pt}
\renewcommand{\arraystretch}{0.82}
\caption{\textbf{Training and evaluation data.} Clip and frame counts are read from the dataset manifests. For the video sources one clip is one full-length recording, not an $81$-frame window. FreiHAND and RHD are static image sources, so their clip column counts images. ``Eval'' is the number of $81$-frame test segments shared with every baseline. For HOT3D and OakInk2 these segments cover only a subset of the available test video. ``Wt.'' is the per-batch sampling weight in percent and is identical for both configurations. HOI4D is held out of training.}
\label{tab:mix}
\begin{tabular}{@{}l rr rr r r@{}}
\toprule
 & \multicolumn{2}{c}{Train} & \multicolumn{2}{c}{Test} & & \\
\cmidrule(lr){2-3}\cmidrule(lr){4-5}
Source & clips & frames & clips & frames & Eval & Wt. \\
\midrule
ARCTIC & 267 & 184{,}373 & 34 & 24{,}863 & 291 & 14 \\
HOT3D & 126 & 444{,}649 & 72 & 256{,}870 & 437 & 18 \\
H2O & 114 & 68{,}929 & 46 & 30{,}724 & 355 & 10 \\
OakInk2 & 550 & 935{,}260 & 77 & 69{,}371 & 202 & 14 \\
Re:InterHand & 43 & 21{,}315 & -- & -- & -- & 16 \\
FreiHAND & 32{,}560 & 162{,}800 & -- & -- & -- & 14 \\
RHD & 41{,}251 & 206{,}255 & -- & -- & -- & 14 \\
\midrule
HOI4D (held out) & -- & -- & 166 & 49{,}800 & 498 & 0 \\
\bottomrule
\end{tabular}
\end{table}

\subsection{Baseline Reproduction}
Every number we compute is produced by a single evaluation protocol applied to per-segment prediction files, and the values quoted from prior work follow that same protocol, so no metric definition changes between the rows of a table. That protocol is the one introduced by ViDiHand~\citep{vidihand2026}, whose metric definitions, on-screen gate, and test segments we adopt unchanged.

Nine of the twelve rows in each block of Table~\ref{tab:main} report methods that ViDiHand also evaluates, and for those the table quotes the published values. EgoForce is not among them, so its row comes from our own run under the same protocol, as do the two \method{} rows and every row of this appendix. Quoting across papers is only sound if the two implementations of the protocol agree, so we did not take that on trust. We re-ran the eight methods with public releases through our own evaluator, and for ViDiHand, which has no public release, we scored predictions that its authors produced at our request on the same test segments, comparing each outcome with the published number. The agreement is close enough for us to treat the two implementations as one protocol, and where a cell differs, the table keeps the published value rather than ours, so that the nine quoted rows remain consistent with the source they come from. The paragraphs below describe how those reproduction runs were set up. Except for ViDiHand, baseline predictions come from the official public release of each method, run without retraining and without edits to its model code, then converted to the shared MANO format before scoring. 

HaMeR~\citep{hamer2024} and Hamba~\citep{hamba2024} use the released \texttt{hamer.\allowbreak ckpt} and \texttt{hamba.\allowbreak ckpt} behind the Detectron2 ViTDet-H detector and the ViTPose+-Huge whole-body keypoint model that produce their hand boxes. WildHands~\citep{wildhands2024} uses \texttt{wildhands.\allowbreak ckpt} and OmniHands~\citep{omnihands2025} its nine-frame video model \texttt{Demo\_\allowbreak Video.\allowbreak pth} with \texttt{config\_\allowbreak video.\allowbreak yaml}, both behind the same detector front end, that is, ViTDet-H plus ViTPose+-Huge. InterWild~\citep{interwild2023} uses the egocentric release \texttt{snapshot\_\allowbreak 6\_\allowbreak ego.\allowbreak pth} behind that front end as well, with its internal BoxNet bypassed at run time so that the same external hand boxes are injected, because that BoxNet mislocalizes hands on egocentric frames. WiLoR~\citep{wilor2025} uses \texttt{wilor\_\allowbreak final.\allowbreak ckpt} with its own YOLO detector, HaWoR~\citep{hawor2025} uses \texttt{hawor.\allowbreak ckpt}, and EgoForce~\citep{egoforce2026} uses its released \texttt{model\_\allowbreak weights.\allowbreak pth}. Dyn-HaMR~\citep{dynhamr2025} runs the default optimization schedule of that repository on top of the HaMeR front end. Because every baseline is scored without retraining on our splits, the training data behind the baseline rows is whatever each model was trained on, and we do not control for that.

Three adaptations are worth stating. Two favor the baseline. WildHands and HaWoR, which take camera geometry as an explicit input, receive the ground-truth intrinsics of the clip, rescaled together with the image where the method requires its own input resolution. Dyn-HaMR receives the ground-truth extrinsics instead of running DROID-SLAM, which removes a failure source unrelated to hand pose. The third works against the baseline, and we state this as a cost to HaWoR. HaWoR's SLAM and infilling stages are skipped and its camera-space motion estimate is scored directly, because the protocol is camera-space and DROID-SLAM is unstable on the $81$-frame ($2.7$\,s) test segments. The two stages are coupled in the released code, since the infiller runs in world space on the SLAM trajectory, so dropping SLAM drops infilling with it. That removes the module meant to carry a hand through frames where it is not detected, so the HaWoR rows report the output of its camera-space hand-motion estimator rather than of its full published pipeline. The $K$-free configuration of \method{} receives no camera information as input. Ground-truth intrinsics enter its evaluation only where the metric definitions above say so, namely to re-project its 3D joints when EPE\textsubscript{2D}-p is scored.

\section{Implementation Details}
\label{sec:supp_impl}
\subsection{Architecture}
The spatial PE $P^{\mathrm{sp}}$ is a learned $16\times16$ grid bilinearly resized to the token resolution, and the ray PE Fourier-encodes each ray's azimuth and elevation with sines and cosines at eight doubling frequencies, after which a zero-initialized MLP maps the resulting features to the decoder width, giving a smooth start to training. In the mixed-PnP solve, a joint votes ($m_j=1$) if it lies at least $5$\,cm in front of the camera and its anchor lies inside the frame by a margin of at least $2\%$ of the image size. When fewer than six joints vote, or when the refit RMS anchor residual exceeds the larger of $15$\,px and a quarter of the hand's 2D bounding-box diagonal, the wrist is placed on its own inverse-projected ray at depth $\hat t_z$. The $K$-free camera fit is a closed-form, differentiable per-axis linear regression over the token grid, guarded by a variance floor of $10^{-4}$ and a bracket on the fitted focal length; a clip that fails either guard, including the fisheye Re:InterHand training arm, falls back to reading the ray field at the anchors, and both decode paths, the per-joint anchor bearings and the wrist fallback, use the fitted camera. The backbone is the Wan2.2-Fun-5B-Control release~\citep{wan2025} distributed with VideoX-Fun, loaded as its low-noise DiT submodel with $30$ blocks of width $3072$, feed-forward width $14336$, $148$ input latent channels, and $48$ output channels, and paired with the Wan 2.2 VAE. We keep the released input and output channel counts unchanged. LoRA adapters of rank $64$ with $\alpha=64$ and no dropout are injected into all ten linear layers of every block, namely the query, key, value, and output projections of both self-attention and cross-attention plus the two feed-forward layers. Adapters are matched by module name rather than by depth, so one set is instantiated in each of the $30$ blocks, giving $5.37$M adapter parameters per block and $161.219$M in total. Three counts are worth keeping apart. Of the $5.002$B pretrained weights the released DiT holds, only the $1.822$M patch embedding is ever updated, and the $30$ transformer blocks stay frozen throughout. With the adapters attached, the instantiated backbone holds $5.16$B parameters. The optimizer is then handed $183.99$M parameters, split by module into $161.219$M in the LoRA adapters, $20.347$M in the decoder and its readout heads, $1.822$M in the patch embedding, $0.596$M in the DiT's own diffusion output head, and $9{,}219$ in the Ray Head. These fall into the three learning-rate groups of the next subsection: the LoRA group, the patch-embedding group, and one group holding the decoder, its readout heads, the Ray Head, and the diffusion output head. Since the forward pass stops at the tap, the parameters a gradient can actually reach are the $85.983$M of LoRA inside the executed blocks, the $1.822$M patch embedding, the $12.287$M of decoder parameters that this configuration routes through, and the Ray Head, for $100.10$M in all. The adapters in the bypassed blocks and the diffusion output head sit on the skipped path and therefore never leave their initialization. We confirmed this on the released checkpoint: after $20$k steps every LoRA $B$ matrix past the tap is still exactly zero, which makes those adapters exact identity maps, and every tensor of the diffusion output head is bit-identical to the pretrained release. The decoder parameters outside the $12.287$M belong to readout variants that this configuration does not select, and we keep them instantiated so that a single checkpoint schema covers every ablation.

\subsection{Optimization}
Only the $183.99$M parameters listed above are registered with the optimizer. The decoder, the Ray Head, and the LoRA adapters are trained from scratch, with the Ray Head and the LoRA $B$ matrices starting at zero so that the network begins the run as the unmodified backbone, while the patch embedding is fine-tuned from its released weights. We run $20$k AdamW steps (weight decay $10^{-2}$, gradient clip $1.0$, cosine decay, $200$ warmup steps) at three learning rates: $2{\times}10^{-4}$ for the decoder and heads, $1{\times}10^{-4}$ for the LoRA adapters, $2{\times}10^{-5}$ for the patch embedding. Each GPU holds four clips, where a clip is $81$ frames for the video datasets and five frames for the two image datasets. The standard configuration runs on $16$ A100 GPUs, for an effective batch of $64$ clips, and the $K$-free configuration on $8$, for an effective batch of $32$. The ablations of Table~\ref{tab:decoder} follow the $K$-free configuration, except that the hand-pooled-query and absolute-PE variants run on half as many GPUs at the same step count and the same number of clips per GPU, the halved-effective-batch caveat noted alongside Table~\ref{tab:decoder}.

\subsection{Loss Weights}
\method{}'s trainable components, namely the LoRA adapters, the patch embedding, the Ray Head, and the Bidirectional Spatiotemporal Decoder, are optimized jointly under Eq.~\eqref{eq:loss}, $\mathcal{L}=\mathcal{L}_{\mathrm{rot}}+\mathcal{L}_{\mathrm{joint}}+\mathcal{L}_{\mathrm{img}}+\mathcal{L}_{\mathrm{cam}}+\mathcal{L}_{\mathrm{pres}}+\mathcal{L}_{\mathrm{tmp}}+\mathcal{L}_{\mathrm{ray}}$. The per-term weights are as follows.

$\mathcal{L}_{\mathrm{rot}}$ supervises orientation and articulation with the geodesic distance $\arccos\big(\tfrac{1}{2}(\mathrm{tr}(\hat R^{\top}R)-1)\big)$ to the ground-truth rotation $R$, plus a rotation-matrix MSE (weight 1 each), and shape with an $\ell_1$ loss (0.1). $\mathcal{L}_{\mathrm{joint}}$ places $\ell_1$ losses on root-relative (weight 10, the dominant term), camera-frame (5), and wrist (2) 3D joints. $\mathcal{L}_{\mathrm{img}}$ supervises, under the \emph{training} camera, the grounded soft-argmax 2D anchors and the re-projected MANO joints (weight 1, and 0.5 for the wrist). $\mathcal{L}_{\mathrm{cam}}$ is an $\ell_1$ loss on the assembled translation (weight 1), with the gradient flowing through the mixed-PnP solve. $\mathcal{L}_{\mathrm{pres}}$ applies binary cross-entropy to existence and visibility (0.5/0.25). $\mathcal{L}_{\mathrm{tmp}}$ penalizes the acceleration of the predicted 3D joints $\hat J_t$, $\|\hat J_{t+1}-2\hat J_t+\hat J_{t-1}\|_1$ (0.5). $\mathcal{L}_{\mathrm{ray}}$ carries weight 1, and the $K$-free configuration's $\mathcal{L}_{\mathrm{fit}}$ carries weight 5 with a linear warmup over the first $500$ steps.

\begin{table*}[t]
\centering
\setlength{\tabcolsep}{2pt}
\renewcommand{\arraystretch}{0.95}
\caption{\textbf{Results on H2O and OakInk2.} Same protocol as Table~\ref{tab:main}. Best per column in bold. MPJPE\textsuperscript{+OOS} needs out-of-sight ground truth, which H2O lacks (--). OakInk2 uses the $202$-segment subset shared with the baselines. $^{\ddagger}$EgoForce post-filters translation with a causal Kalman filter.}
\label{tab:supp_extra}
\begin{tabular}{@{}ll ccc ccc ccc c@{}}
\toprule
& & \multicolumn{3}{c}{Detection} & \multicolumn{3}{c}{3D Pose} & \multicolumn{3}{c}{Orient.\ \& Position} & Temporal \\
\cmidrule(lr){3-5}\cmidrule(lr){6-8}\cmidrule(lr){9-11}\cmidrule(lr){12-12}
& Method & FAcc$\uparrow$ & Recall$\uparrow$ & F1$\uparrow$ & MPJPE-p$\downarrow$ & PA-p$\downarrow$ & MPJPE\textsuperscript{+OOS}$\downarrow$ & EPE\textsubscript{2D}-p$\downarrow$ & GO-p$\downarrow$ & CT-p$\downarrow$ & Jitter$\downarrow$ \\
\midrule
\multirow{12}{*}{\rotatebox{90}{H2O}}
 & InterWild & 0.981 & 0.990 & 0.994 & 21.526 & 8.979 & -- & 19.929 & 17.695 & 0.037 & 16.153 \\
 & HaMeR & 0.980 & 0.990 & 0.992 & 20.079 & 7.221 & -- & 21.786 & 17.971 & 0.034 & 9.307 \\
 & Hamba & 0.960 & 0.979 & 0.987 & 21.412 & 8.390 & -- & 32.978 & 19.510 & 0.038 & 8.285 \\
 & WildHands & 0.989 & 0.995 & 0.996 & 32.587 & 10.508 & -- & 26.297 & 22.033 & 0.082 & 10.894 \\
 & WiLoR & 0.994 & 0.998 & 0.998 & 16.390 & 5.633 & -- & 14.497 & 14.495 & \best{0.023} & 6.117 \\
 & EgoForce$^{\ddagger}$ & 0.990 & 0.996 & 0.997 & 16.219 & 6.020 & -- & 15.849 & 10.416 & 0.024 & 8.046 \\
 & OmniHands & 0.974 & 0.986 & 0.990 & 19.835 & 7.417 & -- & 25.665 & 18.032 & 0.038 & 6.321 \\
 & Dyn-HaMR & 0.988 & 0.999 & 0.997 & 17.213 & 6.390 & -- & 10.790 & 11.057 & 0.030 & 5.349 \\
 & HaWoR & 0.946 & 0.972 & 0.985 & 23.504 & 9.570 & -- & 39.735 & 17.080 & 0.045 & 14.162 \\
 & ViDiHand & 0.997 & 0.999 & 0.999 & 17.320 & 7.987 & -- & 16.544 & 12.159 & 0.040 & 1.990 \\
 & \method{} & \best{0.999} & \best{1.000} & \best{1.000} & \best{9.223} & \best{4.894} & -- & \best{6.088} & \best{5.689} & 0.031 & \best{0.707} \\
 & \method{} ($K$-free) & \best{0.999} & \best{1.000} & \best{1.000} & 10.310 & 5.522 & -- & 9.464 & 6.064 & 0.024 & 0.741 \\
\midrule
\multirow{12}{*}{\rotatebox{90}{OakInk2}}
 & InterWild & 0.547 & 0.722 & 0.819 & 51.013 & 37.038 & 65.619 & 244.590 & 54.710 & 0.179 & 43.499 \\
 & HaMeR & 0.680 & 0.799 & 0.866 & 43.804 & 28.726 & 59.471 & 171.887 & 46.340 & 0.148 & 19.510 \\
 & Hamba & 0.634 & 0.755 & 0.840 & 46.517 & 32.488 & 61.863 & 213.576 & 51.295 & 0.159 & 13.811 \\
 & WildHands & 0.731 & 0.837 & 0.890 & 44.087 & 27.115 & 58.366 & 151.305 & 43.806 & 0.146 & 15.156 \\
 & WiLoR & 0.921 & 0.955 & 0.973 & 26.520 & 12.481 & 45.215 & 39.928 & 25.084 & 0.085 & 9.522 \\
 & EgoForce$^{\ddagger}$ & 0.846 & 0.912 & 0.949 & 36.725 & 18.198 & 55.095 & 81.356 & 35.671 & 0.101 & 30.476 \\
 & OmniHands & 0.530 & 0.688 & 0.787 & 54.712 & 40.373 & 68.024 & 283.245 & 60.157 & 0.174 & 22.562 \\
 & Dyn-HaMR & 0.882 & 0.941 & 0.947 & 28.781 & 14.945 & 47.434 & 46.902 & 25.711 & 0.102 & 7.311 \\
 & HaWoR & 0.818 & 0.891 & 0.935 & 35.428 & 20.670 & 52.775 & 96.442 & 32.043 & 0.120 & 13.780 \\
 & ViDiHand & 0.813 & 0.886 & 0.937 & 38.996 & 22.827 & 56.047 & 101.568 & 38.119 & 0.103 & 3.728 \\
 & \method{} & \best{0.977} & \best{0.985} & \best{0.993} & \best{8.988} & \best{5.887} & \best{8.272} & \best{11.129} & \best{7.222} & \best{0.018} & \best{1.089} \\
 & \method{} ($K$-free) & \best{0.977} & \best{0.985} & \best{0.993} & 9.813 & 6.362 & 9.308 & 12.046 & 7.366 & \best{0.018} & 1.150 \\
\bottomrule
\end{tabular}
\end{table*}

\section{Additional Quantitative Results}
\label{sec:supp_results}
\subsection{Additional In-Domain Datasets: H2O and OakInk2}
\label{sec:supp_extra}
The main comparison (Table~\ref{tab:main}) reports ARCTIC, HOT3D, and held-out HOI4D. Table~\ref{tab:supp_extra} gives the full per-dataset results on two further in-domain datasets, H2O and OakInk2, scored with the same shared evaluator. \method{} again leads every accuracy metric but one, cutting the best baseline MPJPE-p by $66\%$ on OakInk2 ($26.520{\to}8.988$) and by $43\%$ on H2O ($16.219{\to}9.223$). The $K$-free configuration matches the standard configuration's detection on both datasets, with identical detection columns on OakInk2, and stays within $1.1$\,mm of its MPJPE-p ($10.31$ vs.\ $9.22$ on H2O and $9.81$ vs.\ $8.99$ on OakInk2). The only cell a baseline wins is H2O CT-p, where the weak-perspective solve of WiLoR is marginally tighter ($0.023$ vs.\ $0.031$). Two caveats apply. First, each \method{} entry comes from a single training run. Second, both datasets are in-domain for us and out-of-domain for every baseline, so these two blocks read as an in-domain upper bound rather than as the like-for-like comparison ARCTIC and HOT3D provide. On H2O, whose labels are partially pseudo-ground-truth, our model has additionally trained on the distribution it is scored against, whereas OakInk2 carries native MANO ground truth.

\begin{table*}[t]
\centering
\setlength{\tabcolsep}{4pt}
\renewcommand{\arraystretch}{0.95}
\caption{\textbf{Out-of-sight stratification of the wrist-aligned error.} The same wrist-aligned, ground-truth-gated error over three populations: the hand-frames that pass the on-screen gate, the hand-frames that fail it (MPJPE\textsuperscript{OOS}), and all of them together (MPJPE\textsuperscript{+OOS}). Every row satisfies Eq.~\eqref{eq:oosmix} at the counts in the header. None carries the \mbox{-p} penalty, so the in-view column is close to but not the same as the MPJPE-p of Tables~\ref{tab:main} and~\ref{tab:supp_extra}. HOI4D and H2O carry no out-of-sight ground truth. Lower is better. $^{\ddagger}$EgoForce post-filters translation with a causal Kalman filter.}
\label{tab:oos}
\resizebox{\textwidth}{!}{%
\begin{tabular}{@{}l ccc ccc ccc@{}}
\toprule
& \multicolumn{3}{c}{ARCTIC ($43{,}893$ IV / $3{,}247$ OOS)} & \multicolumn{3}{c}{HOT3D ($57{,}985$ / $12{,}394$)} & \multicolumn{3}{c}{OakInk2 ($27{,}988$ / $4{,}736$)} \\
\cmidrule(lr){2-4}\cmidrule(lr){5-7}\cmidrule(lr){8-10}
Method & in view & out of sight & MPJPE\textsuperscript{+OOS} & in view & out of sight & MPJPE\textsuperscript{+OOS} & in view & out of sight & MPJPE\textsuperscript{+OOS} \\
\midrule
 InterWild                & 32.000 & 139.937 & 39.435 & 77.866 & 142.326 & 89.218 & 49.364 & 161.681 & 65.619 \\
 HaMeR                    & 30.576 & 141.005 & 38.183 & 67.956 & 137.847 & 80.264 & 42.361 & 160.588 & 59.471 \\
 Hamba                    & 32.513 & 141.774 & 40.039 & 71.277 & 140.333 & 83.438 & 45.119 & 160.812 & 61.863 \\
 WildHands                & 26.404 & 135.437 & 33.915 & 46.412 & 126.358 & 60.491 & 41.184 & 159.901 & 58.366 \\
 WiLoR                    & 22.775 & 149.477 & 31.502 & 30.921 & 150.697 & 52.014 & 25.428 & 162.143 & 45.215 \\
 EgoForce$^{\ddagger}$    & 23.533 & 149.264 & 32.193 & 44.225 & 151.322 & 63.085 & 36.690 & 163.858 & 55.095 \\
 OmniHands                & 30.706 & 139.371 & 38.191 & 61.152 & 131.289 & 73.503 & 52.382 & 160.461 & 68.024 \\
 Dyn-HaMR                 & 29.084 & 150.464 & 37.445 & 67.786 & 142.547 & 80.952 & 28.127 & 161.532 & 47.434 \\
 HaWoR                    & 46.406 & 148.565 & 53.443 & 70.767 & 150.074 & 84.733 & 34.343 & 161.700 & 52.775 \\
 ViDiHand                 & 22.311 & 149.113 & 31.045 & 21.513 & 151.703 & 44.440 & 38.109 & 162.053 & 56.047 \\
 \method{}                & 15.423 & 35.165 & 16.783 & \best{12.703} & \best{38.649} & \best{17.273} & \best{7.449} & \best{13.131} & \best{8.272} \\
\bottomrule
\end{tabular}}
\end{table*}

\subsection{Out-of-Sight Stratification}
\label{sec:supp_oos}
Table~\ref{tab:oos} decomposes the wrist-aligned error into its in-view and out-of-sight strata for every method on the three datasets that carry out-of-sight ground truth. The table exists so that the two out-of-sight numbers of the main text can be told apart and audited separately, since the blended metric alone cannot distinguish a method that reconstructs out-of-sight hands from one that is merely accurate while the hands are in view.

The decomposition makes the difference concrete. On ARCTIC, ViDiHand and \method{} differ by under $7$\,mm in view ($22.311$ against $15.423$) but by $113.9$\,mm out of sight ($149.113$ against $35.165$). Because only $6.9\%$ of ARCTIC hand-frames are out of sight, the blended column compresses those two very different gaps into $31.045$ against $16.783$. In HOT3D, $17.6\%$ of hand-frames are out of sight, so more of the out-of-sight gap survives into the blend. The blended reduction is therefore larger on HOT3D ($61.1\%$) than on ARCTIC ($45.9\%$), while the stratum reduction runs the other way ($74.5\%$ against $76.4\%$).

Two further points are visible only in the stratum column. First, the baselines are not merely worse out of sight. They collapse into one narrow band far above their in-view errors: all ten sit between $126$ and $164$\,mm on all three datasets, with no ordering that resembles their in-view ranking. This is what one would expect of methods that, in the configuration scored here, have no mechanism for producing a hand pose when no hand appears in the frame. Second, and as a consequence, the strongest baseline out of sight is not the strongest baseline in view: WildHands leads the out-of-sight stratum on ARCTIC ($135.437$) and HOT3D ($126.358$) despite trailing WiLoR and ViDiHand in view, and \method{} still reduces WildHands' error by $74.0\%$ and $69.4\%$. We read the out-of-sight margin as evidence of temporal reconstruction rather than of better perception, and we caution that it is a wrist-aligned margin: Appendix~\ref{sec:supp_limits} states what that margin does not establish.

\subsection{Matched-Detection Errors}
Restricting the evaluation to matched detections separates the placement cost that arises at the image periphery without the camera fit from articulation accuracy. This defines a third population, narrower than the in-view stratum of Table~\ref{tab:oos} because it drops the hand-frames a method failed to detect. The matched-detection numbers are therefore near, but not equal to, the in-view column of that table. On HOT3D the variant without the fit ties the standard configuration once false negatives are excluded ($12.75$\,mm against $12.68$\,mm, wrist-aligned), whereas its coverage-penalized gap in the same order is $18.04$ against $12.89$\,mm. On OakInk2 the two remain close, $7.96$\,mm against $7.47$\,mm. That penalized gap therefore reflects lost coverage at the image periphery, not a loss of hand pose, and the camera fit recovers exactly that coverage (Table~\ref{tab:main}).

\begin{figure}[t]
\centering
\includegraphics[width=\columnwidth]{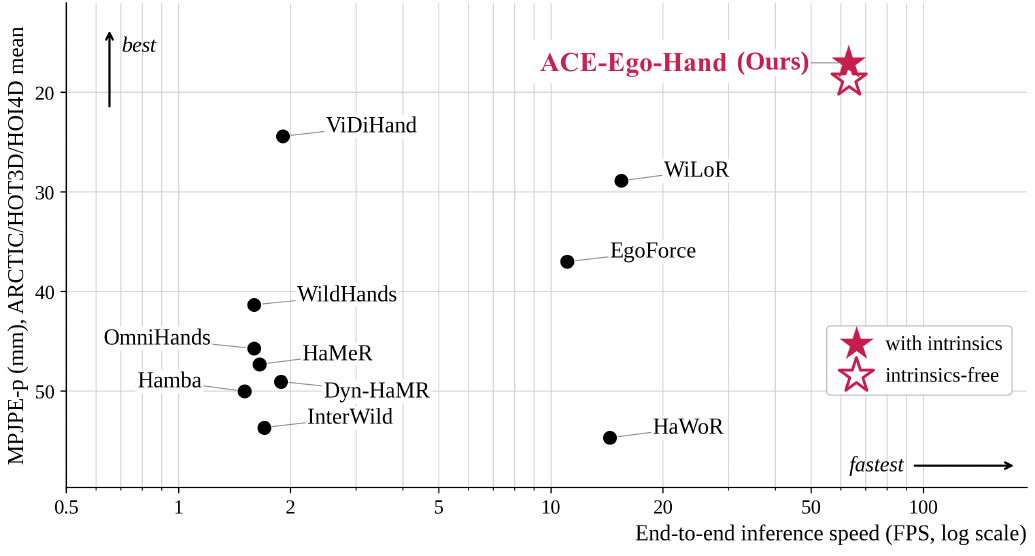}
\caption{\textbf{Accuracy against speed.} We plot MPJPE-p averaged over ARCTIC, HOT3D, and HOI4D against end-to-end clip throughput on one A100, excluding model loading, video decoding, and rendering. Both \method{} configurations run at the same speed.}
\label{fig:statics}
\end{figure}

\subsection{Inference Throughput}
\label{sec:supp_speed}
Figure~\ref{fig:statics} places every method on the accuracy--speed plane. This section states the protocol behind that figure. All methods are timed on the same $81$-frame ARCTIC clip on one A100, from decoded frames in host memory to MANO parameters back on the host, excluding model loading, warmup, rendering, and file I/O. We report the median of at least three passes after a full-clip warmup, and we count the hands actually produced per frame so that missed detections cannot masquerade as speed. Several points depart from that protocol. HaMeR and WildHands warm up on five frames instead of a full clip. The timing for HaMeR is the mean of two passes rather than the median of at least three, and its frame decode sits inside the timed region rather than outside it. That decode costs a few tenths of a second out of $49.0$\,s, so its throughput is understated by well under one percent. Hamba is timed without the compiled selective-scan kernel, which is absent from our environment, so the model falls back to an uncompiled scan. That $1.5$\,fps is therefore a lower bound on what the method can reach. HaWoR is timed in the SLAM-free and infilling-free configuration that produced its accuracy row (Appendix~\ref{sec:supp_protocol}), so its figure does not describe its full published pipeline, which would be slower. The timing for EgoForce is inherited from a separate benchmarking run rather than measured with our own harness, and was taken in fp16 on a matched egocentric $81$-frame clip rather than on our protocol clip. Both departures favor EgoForce. For ViDiHand we use its published accuracy row. Each baseline is timed in the configuration that produced its accuracy row in Table~\ref{tab:main}. \method{} runs at $63.1$\,fps ($63.3$\,fps in the $K$-free configuration) in a single deterministic pass whose runtime is $72\%$ VAE encode, against $1.91$\,fps for ViDiHand, a $33\times$ gap. The five crop-based baselines that share a ViTDet-H$+$ViTPose$+$ front end cluster at $1.5$--$1.7$\,fps, so their cost is set by the detector, not by the reconstruction network. The fastest crop-based method, WiLoR, reaches $15.5$\,fps.

\begin{table}[!t]
\centering
\small
\setlength{\tabcolsep}{1.6pt}
\renewcommand{\arraystretch}{0.82}
\caption{\textbf{Error against distance from the image center.} In-view hand-frames binned by normalized radius $r$, with the share falling in each bin. The columns report the unpenalized in-view stratum, not the matched-detection numbers of Appendix~\ref{sec:supp_results}, so share-weighting a column reproduces the in-view column of Table~\ref{tab:oos}. The two configurations agree to within a millimeter at the center and diverge in translation toward the border, which isolates the cost of sampling bearings from a predicted ray field.}
\label{tab:radius}
\begin{tabular}{@{}l r cc cc@{}}
\toprule
& & \multicolumn{2}{c}{\method{}} & \multicolumn{2}{c}{w/o camera fit} \\
\cmidrule(lr){3-4}\cmidrule(lr){5-6}
Radius $r$ & share & MPJPE & CT & MPJPE & CT \\
\midrule
\multicolumn{6}{@{}l}{\textit{ARCTIC}} \\
\ $0.00$--$0.25$ & $26.1\%$ & 15.81 & 0.019 & 15.49 & 0.019 \\
\ $0.25$--$0.50$ & $57.7\%$ & 14.96 & 0.021 & 14.97 & 0.020 \\
\ $0.50$--$0.75$ & $14.7\%$ & 16.15 & 0.028 & 16.19 & 0.039 \\
\ $>0.75$ & $1.5\%$ & 19.17 & 0.048 & 19.00 & 0.165 \\
\midrule
\multicolumn{6}{@{}l}{\textit{HOT3D}} \\
\ $0.00$--$0.25$ & $7.8\%$ & 12.62 & 0.022 & 12.97 & 0.023 \\
\ $0.25$--$0.50$ & $59.3\%$ & 12.28 & 0.024 & 12.57 & 0.025 \\
\ $0.50$--$0.75$ & $28.5\%$ & 12.68 & 0.025 & 13.12 & 0.058 \\
\ $>0.75$ & $4.4\%$ & 18.80 & 0.035 & 18.19 & 0.190 \\
\bottomrule
\end{tabular}
\end{table}

\section{Analysis of Failure Modes and Design Choices}
\label{sec:supp_analysis}
\subsection{Errors Toward the Image Border}
Without the camera fit of \cref{sec:camera}, $K$-free bearings would be read directly from the predicted ray field, which is defined only over the token grid and becomes unreliable near the image border. This section measures that failure directly. For every in-view hand-frame we compute the normalized distance of its projected center from the principal point, $r=\lVert(u-c_x,v-c_y)\rVert/(\text{diag}/2)$, and bin the errors by $r$ (Table~\ref{tab:radius}). The standard configuration and the variant without the fit are evaluated on identical dumps.

At the image center the two configurations agree to within a millimeter, with translation errors of $0.019$ and $0.019$\,m on ARCTIC and $0.022$ and $0.023$\,m on HOT3D, the standard configuration first. The gap then opens outward, negligibly in the second bin, where the variant without the fit is in fact marginally the tighter of the two on ARCTIC, and sharply in the outer two bins. In the outermost bin its translation error is $0.165$ against $0.048$\,m on ARCTIC, a factor of $3.4$, and $0.190$ against $0.035$\,m on HOT3D, a factor of $5.4$. Wrist-aligned pose is nearly identical between the two configurations in every bin, so the border effect is specific to metric placement rather than to hand pose. This is what the ray-field explanation predicts, because a cell near the frame edge has few neighbors to constrain its direction and the solve inherits that uncertainty as an in-plane shift. HOT3D shows the stronger effect, consistent with its wider field of view placing more hands in the outer bins.

The camera fit removes this failure mode: on an in-view/out-of-sight stratification of the translation error, it reduces the out-of-sight error from $0.55$ to $0.10$\,m on ARCTIC, from $0.52$ to $0.11$\,m on HOT3D, and from $0.45$ to $0.06$\,m on OakInk2, and HOT3D false negatives fall from $2{,}491$ to $74$.

\subsection{Decoded Sequence Length}
\begin{figure}[t]
\centering
\includegraphics[width=\columnwidth]{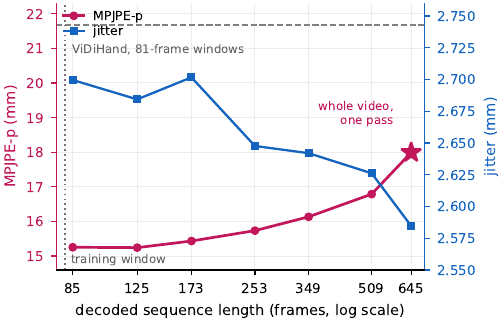}
\caption{\textbf{Effect of decoded sequence length.} We plot \method{} accuracy (MPJPE-p, left axis) and smoothness (Jitter, right axis) against the length decoded in one pass, MPJPE-p in mm and Jitter in mm/frame$^2$. The same $291$ ARCTIC segments are scored at every length. The dotted vertical line marks the $81$-frame training window, the dashed line marks $81$-frame-window MPJPE-p, and the star marks every recording decoded end to end in one pass. The star's abscissa is nominal, at $645$ frames: the recordings themselves run $594$ to $1{,}087$ frames, so the star understates that horizon.}
\label{fig:length}
\end{figure}

Long-horizon decoding is a property of the \method{} architecture rather than a separate mode. This section measures what that costs. All results in the main text are scored on the $81$-frame segments shared with every baseline. Those $81$ frames occupy $21$ latent frames, and a segment whose first frame does not fall on a latent boundary is covered by one further latent frame, so the decoder is run over $21$ or $22$ latent frames depending on where the segment starts. Nothing in the model requires that length: the temporal attention uses rotary \emph{relative} position encoding, so no absolute clip length is baked into the weights and the same checkpoint decodes any length in one pass. Figure~\ref{fig:length} sweeps the length decoded in one pass on the identical $291$ ARCTIC segments. Only the decoded window changes. The scored segments are fixed $81$-frame slices of the $34$ ARCTIC test recordings, which run from $594$ to $1{,}087$ frames. Each window is anchored at the segment it is scored on and clamped to end no later than the recording. The added context is therefore neighboring frames of the same recording, mostly the frames that follow the segment.

Accuracy is flat well past the training window: at $125$ frames, more than $50\%$ longer than that window, MPJPE-p is marginally \emph{better} than at the segment-length setting. Beyond that accuracy degrades gently, reaching $17.99$\,mm when each recording is decoded in a single pass, $7$ to $13\times$ the training window. Jitter, by contrast, falls from $2.700$ to $2.585\,\mathrm{mm}/\mathrm{frame}^2$. The temporal resolution does not change with the pass length, because the VAE always carries four frames per latent. A longer window therefore adds context for the temporal attention to average over without adding any degree of freedom between latent samples. The net effect is $4\%$ additional smoothing at a cost of $18\%$ in MPJPE-p. Even so, \method{} stays $17\%$ below the windowed $21.668$\,mm of ViDiHand at that setting, so whole-recording decoding remains usable when the application needs a single pass. The main text reports the segment-length setting because it is the like-for-like comparison with the baselines.

\subsection{Tap Depth}
The tap sits at block~$15$ of the $30$-block DiT, a choice fixed before the model reported in this paper was trained. This ablation retrains four models that differ only in tap depth and scores all four, to measure what tap depth costs and buys. The four runs share one reduced training configuration, $20$k steps at effective batch $16$ over a mixture of ARCTIC and HOT3D, rather than the seven training sets and the larger batch of the main recipe. They also decode translation by inverse projection rather than by the mixed-PnP solver of the main model. All four are scored at step $20$k on the same $291$-segment ARCTIC test split, so absolute errors are comparable within the sweep but not with Table~\ref{tab:main}. Table~\ref{tab:tap} reports the sweep.

Accuracy has saturated by the middle of the stack. Tapping at block~$10$, which runs $11$ of the $30$ blocks, costs $1.67$\,mm of MPJPE-p and is the only setting that is clearly worse, degrading every column at once. Descending past block~$15$ buys at most $0.17$\,mm while running $21$ and $25$ of the $30$ blocks instead of $16$, and the two deeper taps do not order consistently: block~$20$ is marginally better than block~$24$ on five of the six metrics but loses on global orientation. We therefore read the residual difference as run-to-run noise rather than as a trend. Block~$15$ therefore captures nearly all of the available accuracy at roughly half the forward cost, confirming the choice the main model was trained with.

\begin{table}[!t]
\centering
\small
\setlength{\tabcolsep}{2.5pt}
\renewcommand{\arraystretch}{0.9}
\caption{\textbf{Tap depth.} Four runs differing only in which DiT block is read, scored at step $20$k on the same $291$ ARCTIC segments under the reduced sweep recipe. Block indices are zero-based and a tap returns the output of its block, so a tap at block~$i$ runs the first $i{+}1$ of the $30$ blocks. Each row label gives that executed count. Lower is better throughout.}
\label{tab:tap}
\resizebox{\columnwidth}{!}{%
\begin{tabular}{@{}l cccccc@{}}
\toprule
Tap (blocks run) & MPJPE-p & PA-p & EPE\textsubscript{2D}-p & GO-p & CT-p & Jitter \\
\midrule
Block 10 ($11/30$) & 17.77 & 8.70 & 11.63 & 12.58 & 0.030 & 2.74 \\
\textbf{Block 15} ($\mathbf{16/30}$) & 16.10 & 7.76 & 9.99 & \best{11.98} & 0.028 & 2.66 \\
Block 20 ($21/30$) & \best{15.93} & \best{7.68} & \best{9.87} & 12.15 & \best{0.021} & \best{2.59} \\
Block 24 ($25/30$) & 15.95 & 7.70 & 9.93 & 12.02 & 0.023 & 2.60 \\
\bottomrule
\end{tabular}}
\end{table}

\subsection{Clean versus Noised Latents}
\method{} reads the backbone at $\sigma=0$, where the input latent is the clean VAE code and no noise is added. The main text argues for this on determinism grounds. The measurement that supports that argument is reported here. We retrain with $\sigma=0.5$ in place of $\sigma=0$, so the encoder reads a half-noised latent at the timestep bound to that noise level. The two runs share the training data, the schedule, and the evaluator, are scored on the same $291$ ARCTIC segments, and are read at the same $20$k-step checkpoint (Table~\ref{tab:sigma}).

Reading the clean latent is better on every metric. MPJPE-p improves by $2.40$\,mm ($17.66$ to $15.26$) and translation error by $39\%$ ($0.0343$ to $0.0208$\,m), while the two runs sit within $0.08\,\mathrm{mm}/\mathrm{frame}^2$ of each other on Jitter. The gain is concentrated in metric placement rather than articulation, since PA-p moves by only $0.52$\,mm. Injecting noise degrades geometry that the deterministic read preserves, and the sample diversity it introduces goes unused because no generative pass is run. The comparison covers two noise levels and one seed per level, so it shows that $\sigma{=}0$ is the better of the two settings, not that error varies monotonically with $\sigma$. The gap is not an artifact of the checkpoint we read. Across the $31$ validation checkpoints logged between step $5$k and step $20$k, the noised run is the worse of the two on $29$, by a mean of $1.2$\,mm of wrist-aligned validation error.

\begin{table}[!t]
\centering
\small
\setlength{\tabcolsep}{2pt}
\renewcommand{\arraystretch}{0.82}
\caption{\textbf{Clean versus noised latent readout.} Two runs that differ in the noise level $\sigma$ at which the backbone is read, scored by the shared evaluator on the same $291$ ARCTIC segments at the same $20$k-step checkpoint. Lower is better throughout.}
\label{tab:sigma}
\begin{tabular}{@{}l cccccc@{}}
\toprule
Latent readout & MPJPE-p & PA-p & EPE\textsubscript{2D}-p & GO-p & CT-p & Jitter \\
\midrule
clean, $\sigma{=}0$ & \best{15.26} & \best{7.47} & \best{9.18} & \best{11.81} & \best{0.021} & \best{2.70} \\
noised, $\sigma{=}0.5$ & 17.66 & 7.99 & 9.78 & 12.62 & 0.034 & 2.78 \\
\bottomrule
\end{tabular}
\end{table}

\section{Qualitative Results and Application}
\label{sec:supp_qualapp}
\subsection{Extended Qualitative Comparison}
\label{sec:supp_qual}
\begin{figure*}[p]
\centering
\includegraphics[width=0.915\textwidth]{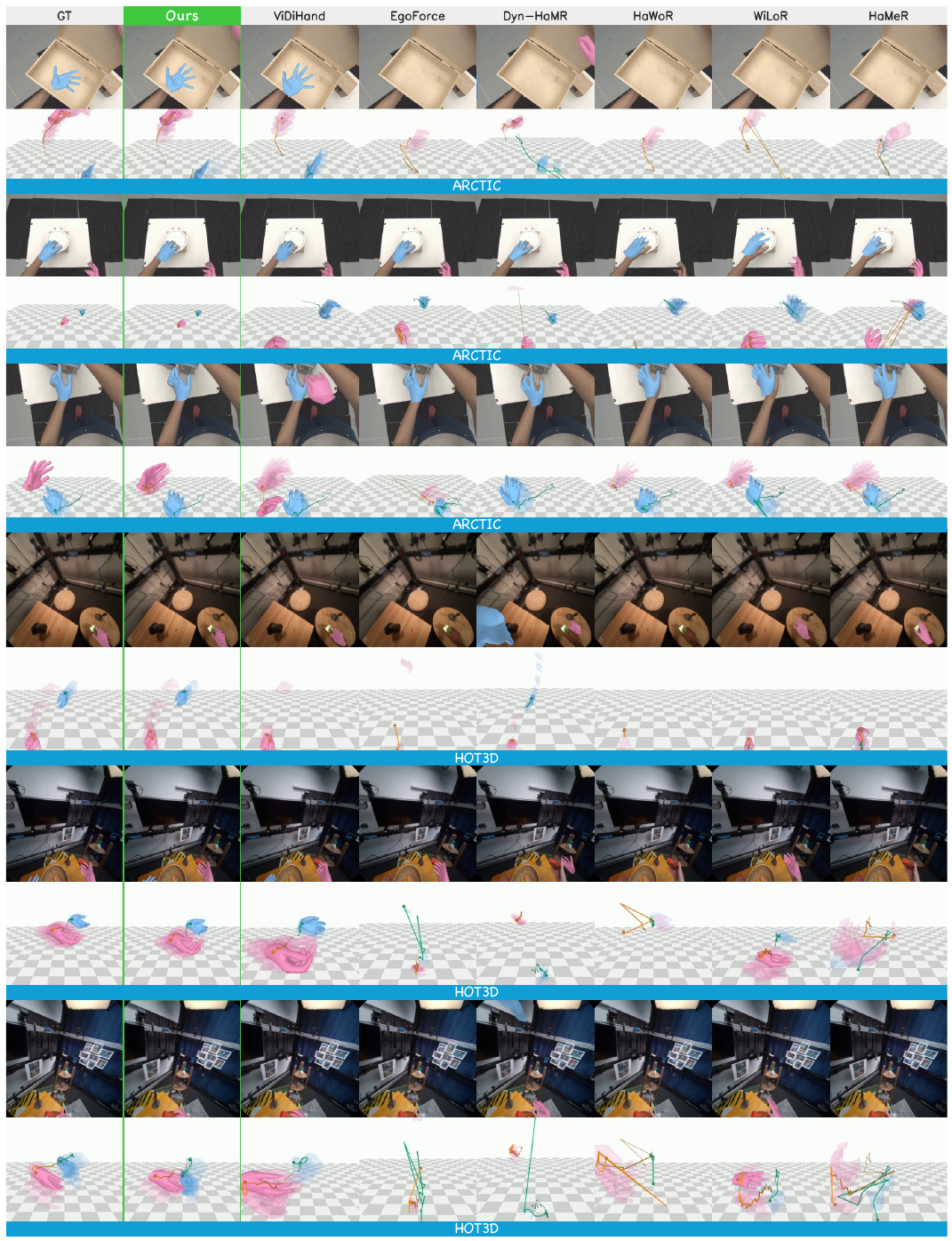}
\caption{\textbf{Extended qualitative comparison on ARCTIC and HOT3D.} Columns: ground truth, \method{}, and six baselines (ViDiHand, EgoForce, Dyn-HaMR, HaWoR, WiLoR, HaMeR). Each dataset block shows a camera-space overlay row above the corresponding 3D reconstruction, which uses the ground-truth camera poses these two datasets provide.}
\label{fig:supp_qual}
\end{figure*}

\begin{figure*}[p]
\centering
\includegraphics[width=0.99\textwidth]{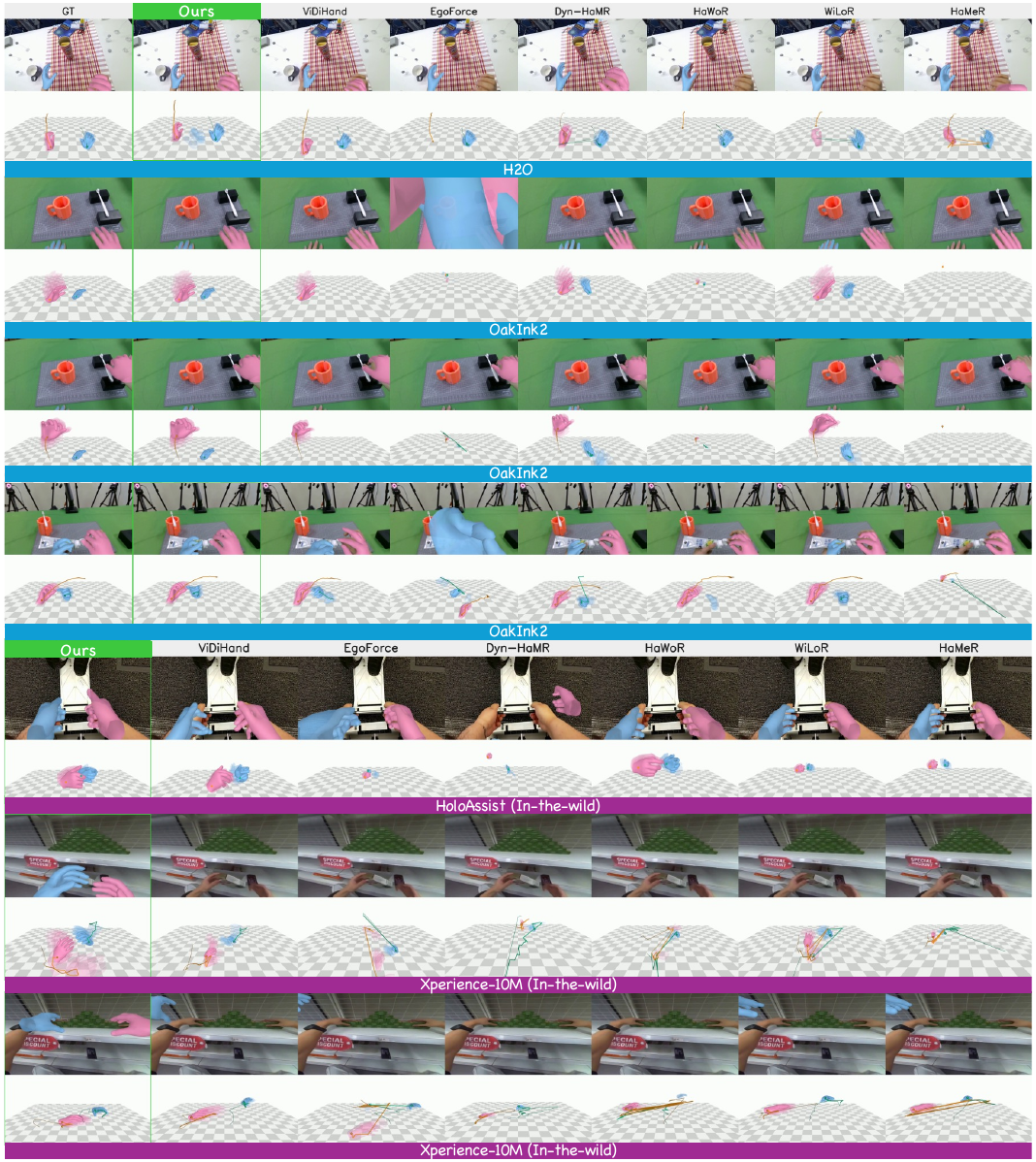}
\caption{\textbf{Extended qualitative comparison on H2O, OakInk2, and in-the-wild HoloAssist and Xperience-10M.} Same columns and same layout as Figure~\ref{fig:supp_qual}. H2O and OakInk2 supply ground-truth camera poses for the 3D row, while the two in-the-wild sources carry no annotation and are shown for qualitative inspection only.}
\label{fig:supp_qual2}
\end{figure*}

Figures~\ref{fig:supp_qual} and~\ref{fig:supp_qual2} extend the comparison of Figure~\ref{fig:compare}. Figure~\ref{fig:supp_qual} covers the two main benchmarks, ARCTIC and HOT3D. Figure~\ref{fig:supp_qual2} covers H2O and OakInk2 together with two in-the-wild egocentric datasets, HoloAssist~\citep{holoassist2023} and Xperience-10M~\citep{xperience10m2026}. Neither carries ground-truth hand annotation here, and neither enters any training mixture or evaluation table in this paper. We include them to place the comparison on footage that none of the methods was tuned on. Each block pairs a camera-space overlay row with the corresponding 3D reconstruction, and the columns hold the ground truth, where it exists, followed by \method{} and six baselines. Every method predicts in the camera frame. For the datasets that provide camera poses, the 3D rows place those camera-frame predictions in a shared world frame, following the same convention as Figure~\ref{fig:compare}. The clips are chosen to contain OOS intervals, because that is where the methods separate most visibly. Single-frame baselines drop the hand entirely once it leaves the frame and re-acquire it in a different pose, while world-space baselines keep producing a hand but drift away from the ground-truth trajectory. \method{} carries the hand through the gap and re-aligns it with the visible evidence on re-entry.

\begin{figure*}[p]
\centering
\includegraphics[width=0.70\textwidth]{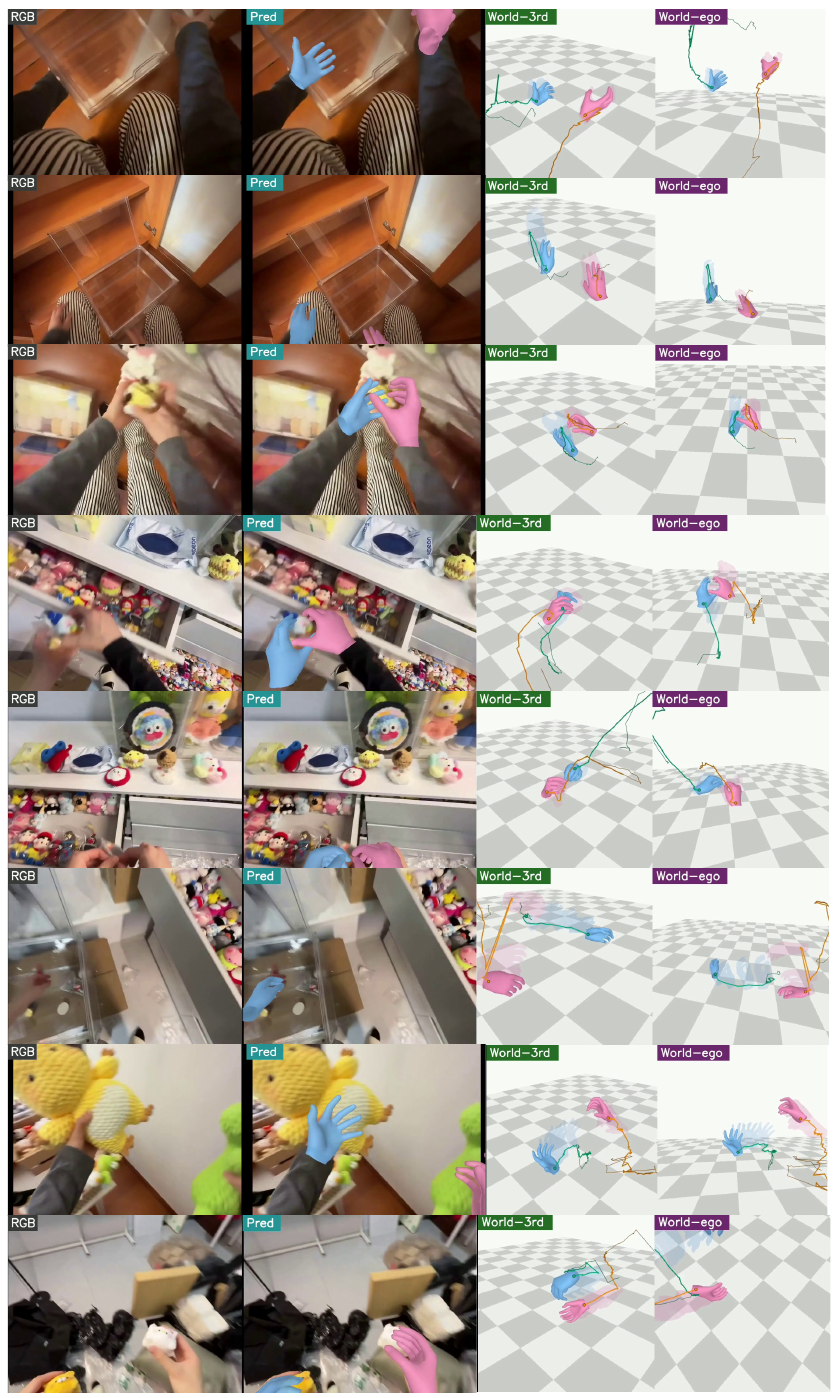}
\caption{\textbf{\method{} qualitative results on in-the-wild videos without ground truth.} Columns: source RGB, predicted hands overlaid on the source video, and the same prediction rendered from a fixed third-person viewpoint and from the head-camera pose.}
\label{fig:supp_itw}
\end{figure*}

\subsection{Qualitative Results Without Ground Truth}
Figure~\ref{fig:supp_itw} shows \method{} at eight moments drawn from two egocentric recordings for which no ground-truth annotation exists, so nothing in the pipeline can be tuned to them. Each row holds the source frame, the camera-space overlay of the prediction, and two renders of the same prediction, one from a fixed third-person viewpoint and one from the head-camera pose. \method{} predicts hands in the camera frame and does not estimate camera motion, and these clips carry no camera annotation, so the two rendered views show the recovered hands themselves rather than a globally referenced trajectory. Without ground truth to overlay, what remains verifiable is whether the two hands stay metrically separated, keep a consistent scale as the head turns, and continue along a plausible path while they are outside the frame. The figure shows these properties, which the benchmark tables measure numerically, on footage that carries no annotation.

\subsection{Retargeting to a Dexterous Hand}
\begin{figure*}[p]
\centering
\includegraphics[width=0.99\textwidth]{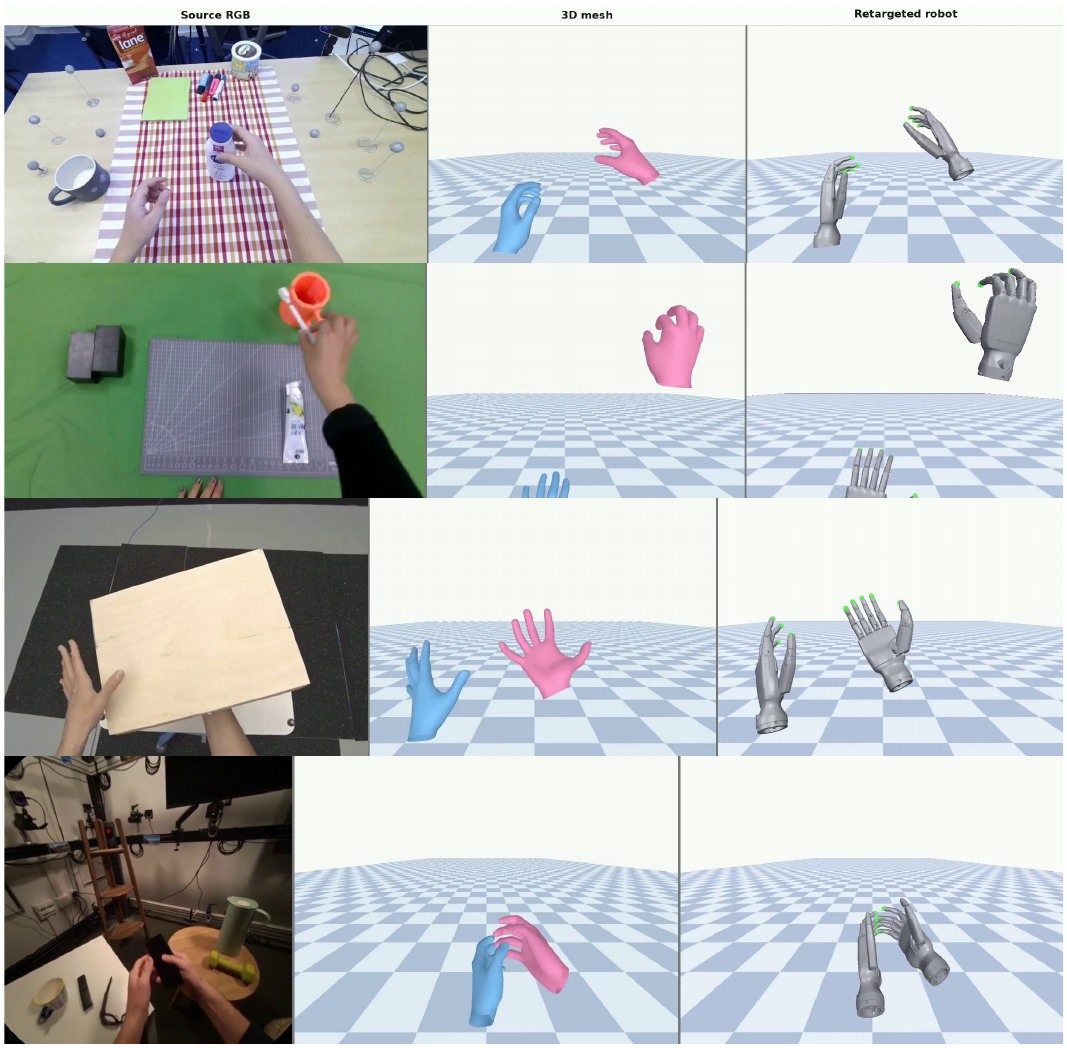}
\caption{\textbf{From egocentric video to a dexterous hand, step by step.} One clip from each of ARCTIC, HOT3D, H2O, and OakInk2. Columns: source RGB, the hand mesh recovered by \method{}, and the trajectory retargeted to an Inspire Hand. Finger flexion is mapped from the predicted 3D joints rather than from the MANO articulation parameters.}
\label{fig:robot}
\end{figure*}

\begin{figure*}[tp]
\centering
\includegraphics[width=0.99\textwidth]{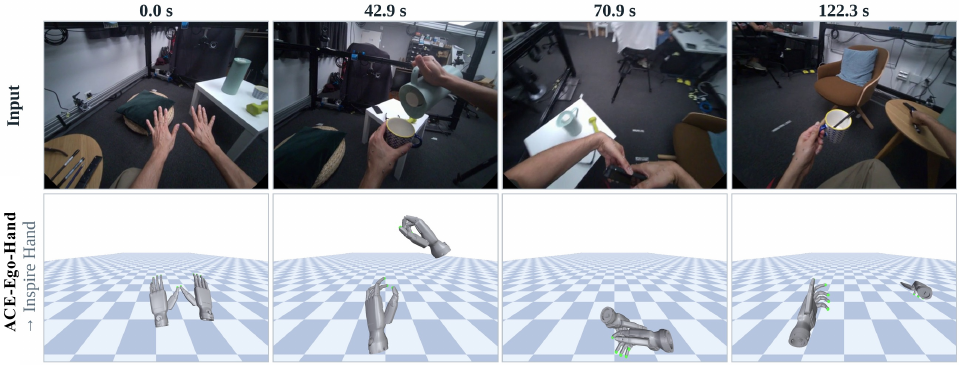}
\caption{\textbf{Retargeting over a two-minute recording.} A single recording sampled at four instants, with the input frame above and the retargeted dexterous hand below. The horizon is roughly $45$ times the $81$-frame training window, and both hands are still tracked and correctly identified at the last of the four instants.}
\label{fig:robot_strip}
\end{figure*}

We show the step before policy learning, qualitatively: replaying the recovered bimanual trajectories on a dexterous Inspire Hand, shown in a neutral gray color scheme for legibility, with each panel rendered under the intrinsics of the source clip so robot and video share one projection. Finger flexion is read off the predicted 3D joints rather than the MANO articulation parameters. Figure~\ref{fig:robot} places the source frame, the recovered hand mesh, and the retargeted robot hand side by side for one clip from each of the four in-domain datasets, so the two conversion steps can be inspected separately. Figure~\ref{fig:robot_strip} then follows a single recording over two minutes, sampled at four instants, to show that the conversion still holds at the end of a horizon far longer than the $81$-frame training window.

A demonstration requires the three properties \method{} provides: metric placement in the camera frame, clip-level shape, and OOS coherence where per-frame methods drop frames. At $63.1$\,fps, the trajectory recovery that feeds this conversion is not the collection bottleneck. Whether such trajectories improve downstream policy learning remains for future work.

\section{Limitations, Failure Cases, and Data Use}
\label{sec:supp_limits}
\subsection{Limitations and Failure Cases}
Five limitations follow from the analyses above. Where possible, we state the concrete failure signature in the units a user would see.

\paragraph{The $K$-free configuration trades a small pose margin for calibration freedom.} Its camera fit keeps detection and metric placement at the level of the calibrated solve (Table~\ref{tab:main}), and the residual cost sits in wrist-aligned pose: MPJPE-p runs $0.6$--$1.4$\,mm above the standard configuration on the in-domain benchmarks, consistent with the $\mathcal{L}_{\mathrm{fit}}$ term competing for shared capacity; annealing its weight is a natural next step. The fit assumes a pinhole camera, so fisheye clips fall back to reading the ray field directly, and a radial bearing model remains future work. The failure signature is a slightly looser articulation, not a lost detection.

\paragraph{Long single-pass decoding trades accuracy for smoothness.} Decoding a whole recording in one pass costs $18\%$ in MPJPE-p relative to the $81$-frame setting ($15.26$ to $17.99$\,mm), so a single long pass is an operating point rather than a free improvement. The failure signature is a trajectory that is smoother but less tightly fitted to the ground truth.

\paragraph{Ground truth and splits are uneven across the five benchmarks.} Only ARCTIC and H2O are subject-disjoint. HOT3D is split at the recording level and OakInk2 at the sequence level, so subject overlap between train and test cannot be excluded there. As noted in Appendix~\ref{sec:supp_protocol}, HOI4D and H2O rely partially on pseudo-ground-truth annotation, which flatters any method that resembles the per-frame estimator behind those labels. On H2O, our own model was trained on that label distribution. The H2O and OakInk2 blocks of Table~\ref{tab:supp_extra} also compare an in-domain \method{} against baselines trained elsewhere, unlike the like-for-like ARCTIC and HOT3D comparison of Table~\ref{tab:main}.

\paragraph{The out-of-sight gain is wrist-aligned and does not certify placement.} Both MPJPE\textsuperscript{OOS} and MPJPE\textsuperscript{+OOS} align the wrist before measuring, so the margins of Table~\ref{tab:oos} certify that the articulation and the global orientation of an out-of-sight hand stay close to the ground truth, and nothing more. Where the hand actually sits while it is out of sight depends on a regressed depth and on a bearing that no image evidence constrains, and on the solver fallback of Appendix~\ref{sec:supp_impl} once too few joint anchors survive inside the frame. We therefore read the out-of-sight result as temporal coherence through the gap rather than as metric localization during the gap, and the failure signature is a plausibly articulated hand on a trajectory that has drifted in depth. No table in this paper measures absolute out-of-sight placement, because none of the five benchmarks scores it.

\paragraph{The method is offline by construction.} What makes \method{} offline is its dependence on clip-level context, not the direction of temporal attention in the decoder. The backbone attends over the whole clip in one pass, so latency is bounded below by the length of the clip rather than by the $63.1$\,fps throughput. \method{} is a tool for curating manipulation data from recorded video, not a closed-loop controller. Three directions follow from that boundary. Distilling the encoder into a causal variant would trade clip-level context for streaming operation and would also quantify how much of the accuracy rests on future frames. Bounding the pass length while keeping identity across passes would remove the dependence of latency on clip length. Carrying hand identity and shape over recordings of several minutes would extend the horizon beyond what one pass now holds.

\subsection{Ethics and Data Use}
All five benchmarks and the three auxiliary training sources are public research datasets, as are the two in-the-wild datasets in Figure~\ref{fig:supp_qual2}. We obtained each through its official channel and use it under the terms of that release. For Xperience-10M those terms are controlled access for non-commercial research. We collected no new human-subject data and we redistribute no dataset. Egocentric recordings can contain identifiable people and private environments, so we use them only to fit hand geometry. The model takes video, together with camera calibration in the standard configuration, and outputs MANO pose, shape, and camera-frame translation. The model neither receives nor produces identity, demographic, or biometric-identification attributes, and is not a person recognizer. Frames shown in figures are taken from those public releases and are reproduced under the same terms, with one exception. The frames in Figure~\ref{fig:supp_itw} come from a licensed egocentric dataset outside those benchmarks, used for qualitative illustration only. That dataset enters no training mixture and no evaluation table. The intended downstream use is imitation learning from recordings whose subjects consented to being recorded. Running the model on third-party video without that consent is a misuse we do not endorse.

\end{document}